\documentclass{article}

\usepackage[final, nonatbib]{neurips_2022}

\usepackage[utf8]{inputenc} 
\usepackage[T1]{fontenc}    
\usepackage{url}            
\usepackage{booktabs}       
\usepackage{amsfonts}       
\usepackage{nicefrac}       
\usepackage{microtype}      
\usepackage{xcolor}         
\usepackage{amsmath}
\usepackage{amssymb}
\usepackage{amsthm}
\usepackage{algorithm}
\usepackage{algorithmic}
\usepackage{listings}
\usepackage[pdftex]{graphicx}
\usepackage{color}
\usepackage{multirow}
\usepackage{float}
\usepackage{color}
\definecolor{citecolor}{HTML}{0071BC}
\definecolor{linkcolor}{HTML}{ED1C24}
\usepackage[pagebackref=false, breaklinks=true, colorlinks, citecolor=citecolor, linkcolor=linkcolor, bookmarks=false]{hyperref}

\pdfoutput=1
\title{LGDN: Language-Guided Denoising Network \\ for Video-Language Modeling}

\author{Haoyu Lu$^{1,2}$~~~Mingyu Ding$^3$~~~Nanyi Fei$^{1,2}$~~~Yuqi Huo$^4$~~~Zhiwu Lu$^{1,2,}$\thanks{The corresponding author.}\\
$^1$Gaoling School of Artificial Intelligence, Renmin University of China, Beijing, China\\
$^2$Beijing Key Laboratory of Big Data Management and Analysis Methods\\
$^3$The University of Hong Kong, Pokfulam, Hong Kong\\
$^4$JD Corporation, Beijing, China\\
{\tt\small \{lhy1998,~luzhiwu\}@ruc.edu.cn}
}

\begin{document}

\maketitle

\begin{abstract}
  Video-language modeling has attracted much attention with the rapid growth of web videos. Most existing methods assume that the video frames and text description are semantically correlated, and focus on video-language modeling at video level. However, this hypothesis often fails for two reasons: (1) With the rich semantics of video contents, it is difficult to cover all frames with a single video-level description; (2) A raw video typically has noisy/meaningless information (e.g., scenery shot, transition or teaser). Although a number of recent works deploy attention mechanism to alleviate this problem, the irrelevant/noisy information still makes it very difficult to address. To overcome such challenge, we thus propose an efficient and effective model, termed Language-Guided Denoising Network (LGDN), for video-language modeling. Different from most existing methods that utilize all extracted video frames, LGDN dynamically filters out the misaligned or redundant frames under the language supervision and obtains only \textbf{\emph{2--4 salient frames}} per video for cross-modal token-level alignment. Extensive experiments on five public datasets show that our LGDN outperforms the state-of-the-arts by large margins. We also provide detailed ablation study to reveal the critical importance of solving the noise issue, in hope of inspiring future video-language work.
\end{abstract}

\section{Introduction}

\begin{figure}[t]
    \centering
    \includegraphics[width=0.99\linewidth]{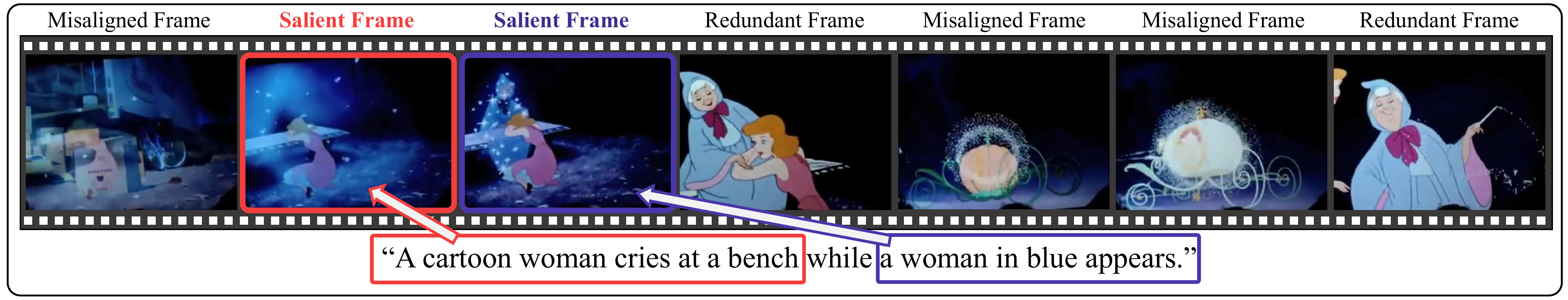}
    \vspace{-0.05in}
      \caption{
      An example of video-text pair where the raw video contains misaligned and redundant video frames given the text description. Instead of aggregating all video frames for token-level alignment, we obtain \emph{only 2--4 salient frames} per video by filtering out the misaligned ones. We find that utilizing 2--4 salient frames is much more effective while enjoying faster speed.}
    \label{fig:teaser}
    \vspace{-0.05in}
\end{figure}

Humans are exposed to the world through a variety of sensory organs, such as eyes, ears, and the sense of touch. In the past few years, multi-modal data (e.g., text or video) has grown and accumulated rapidly on the Internet, which brings the increasing demands for video-language understanding. As one of the fundamental topics, video-language modeling is still challenging due to the heterogeneity of the video-text data. More notably, the video-text data is typically noisy (e.g., misaligned or semi-relevant, as shown in Figure~\ref{fig:teaser}), leading to intractable video-language modeling. 

The dominant paradigm~\cite{dong2018predicting, liu2019use, gabeur2020multi, he2021improving, wang2021t2vlad} for video-language modeling is to first extract language features and dense video features via off-the-shelf language and vision models (e.g., BERT~\cite{jacob2019bert}, 3D CNN~\cite{xie2018rethinking}), and then model the cross-modal representation by defining the objective function (e.g., triplet loss~\cite{hermans2017defense}) within a joint semantic space. Although achieving great success, these methods typically densely sample frames from a full sequence of raw video to obtain richer representation and thus cost excessive computation. Since the heavy computation makes it challenging to train the whole network end-to-end, they often achieve sub-optimal performance in video-language modeling. Recently, ClipBERT~\cite{lei2021less} proposes a sparse sampling strategy to tackle this drawback. Concretely, ClipBERT first samples video frames sparsely (8--16 frames per video), and then models the cross-modal alignment at frame-level. This sparse sampling paradigm enables end-to-end training, leading to much better performance. Nevertheless, token-level cross-modal interaction, which has achieved great success in image-text modeling~\cite{kim2021vilt, li2021albef}, is still not well explored for video-language modeling due to the heavy resource computation (even with 8--16 frames per video).
Moreover, both the dominant paradigm and ClipBERT's sparse sampling paradigm assume that video frames and the text description (w.r.t. a video-text pair) are semantically correlated, which is often invalid in practice. 

The correlation hypothesis often fails for two reasons: (1) With the rich semantics of video contents, it is hard to cover all frames with a single video-level description; (2) A raw video often has noisy or meaningless information (e.g., scenery shot, transition or teaser). For the dominant paradigm which utilizes densely-sampled frames, though often with self-attention mechanism~\cite{vaswani2017transformer}, the irrelevant/noisy information makes it hard to learn high-quality video-language representation. For the sparse sampling paradigm used in ClipBERT that models the cross-modal alignment at frame-level, the misaligned frame-text pairs are wrongly forced to become closer, which inevitably leads to inaccurate cross-modal alignment. Overall, due to this noise issue (see Figure~\ref{fig:teaser}), video-language modeling is still challenging. Note that humans also encounter such problem in reality, but seem to be born with the ability to resist noise. That is, everyone can quickly scan through the entire video, easily ignore the noisy frames and focus on the salient ones given the text.

Motivated by this human ability, we propose a \textbf{L}anguage-\textbf{G}uided \textbf{D}enoising \textbf{N}etwork termed LGDN to dynamically filter out irrelevant or redundant information under the language supervision for better video-language modeling. 
Concretely, we devise a \textbf{S}alient \textbf{F}rame \textbf{P}roposal (SFP) mechanism which adopts four strategies to estimate frame-level relevance scores under the language supervision and proposes/selects only salient frames (per video) for precisely video-language modeling. Although the frame embeddings and text embeddings can be (roughly) aligned by introducing a \textbf{M}omentum \textbf{V}ideo-Level \textbf{C}ontrastive \textbf{L}earning (MVCL) module, it is vital to precisely establish frame-text alignment for proposing salient frames. 
Therefore, based on multiple instance learning (MIL), we propose a \textbf{M}omentum \textbf{F}rame-Level Multiple Salient-instance learning (MSL) -\textbf{C}ontrastive \textbf{L}earning (MFCL) module for video-language modeling at frame-level. Finally, with our SFP mechanism, we propose a \textbf{L}anguage-Guided \textbf{S}alient \textbf{F}rame \textbf{M}atching (LSFM) module for fine-grained alignment, which adopts a token-aware cross-attention Transformer for cross-modal token-level alignment. 

Our main contributions are as follows: (1) We devise a salient frame proposal mechanism that can dynamically filter out irrelevant information under the language supervision, meanwhile maintaining salient information. (2) We propose an end-to-end framework termed LGDN for video-language modeling with cross-modal interaction at three levels: language-guided salient frame matching at token-level, momentum frame-level MSL-contrastive learning, and momentum video-level contrastive learning. (3) We evaluate our LGDN on five public datasets and find that our LGDN outperforms the latest competitors by large margins. We also provide detailed ablation study to reveal the critical importance of solving the noise issue, in hope of inspiring future video-language work.

\section{Related Work}

\paragraph{Video-Language Modeling.} 
Video-language modeling, a fundamental research topic that is beneficial for search engine and video recommendation, has attracted a lot of attention in recent years with the rapid growth of web videos. Previous works have made great efforts to model richer representations for video and text modalities and then align the features of the two modalities by the objective function (e.g., triplet loss). One common representative approach~\cite{chen2020fine, jin2021hierarchical} is to adopt a Graph Convolution Network (GCN) to extract richer information for video-text retrieval. Another representative approach~\cite{liu2019use, gabeur2020multi, he2021improving, yang2021taco, liu2021hit} is to exploit extra experts (e.g., object, motion, speech) for video-language modeling. 
Recently, ClipBERT~\cite{lei2021less} proposes a sparse sampling strategy that enables end-to-end training, thus achieving higher performance. Moreover, Frozen in Time~\cite{bain2021frozen} also follows a sparse sampling paradigm, and proposes an end-to-end trainable model that is designed to take advantage of both large-scale image and video captioning datasets. 
However, as illustrated in Figure~\ref{fig:teaser}, a raw video typically has noisy/meaningless information, and thus the presence of misaligned frames is inevitable during video-language modeling. Note that most existing methods assume that the video frames and paired text are semantically correlated, without considering the noise phenomenon. Although a self-attention mechanism has been widely applied, the misaligned frames still harm the cross-modal alignment. In this work, we thus propose a salient frame proposal mechanism to effectively (and directly) address this problem.

\paragraph{Cross-Modal Alignment Objective Functions}
Most previous methods adopt triplet loss as a major objective function for video-language modeling. CGMSCD~\cite{he2021improving} points out that the triplet loss sometimes leads to a wrong learning direction and thus devises an adaptive margin triplet loss for representation learning. More recent works~\cite{radford2021learning, huo2021wenlan, jia2021scaling} propose to apply the InfoNCE contrastive loss~\cite{wu2018unsupervised, oord2018representation, chen2020simple} to enhance representation learning. Particularly,
BriVL~\cite{huo2021wenlan}, ALBEF~\cite{li2021albef} and COTS~\cite{lu2022cots} introduce a momentum mechanism~\cite{he2020moco} to maintain more negative samples for image-text contrastive learning. Following these state-of-the-art models, we propose momentum video-level contrastive learning for video-text global alignment in this paper. Note that MIL-NCE~\cite{miech2020end} enhances the InfoNCE loss with multiple-instance learning (MIL) to cope with the misaligned narration descriptions in HowTo100M~\cite{miech2019howto100m}. In this work, we thus propose momentum frame-level MSL-contrastive learning to assist in addressing the misaligned frame problem.

\section{Methodology}

\begin{figure*}[t]
    \centering
    \includegraphics[width=0.99\textwidth]{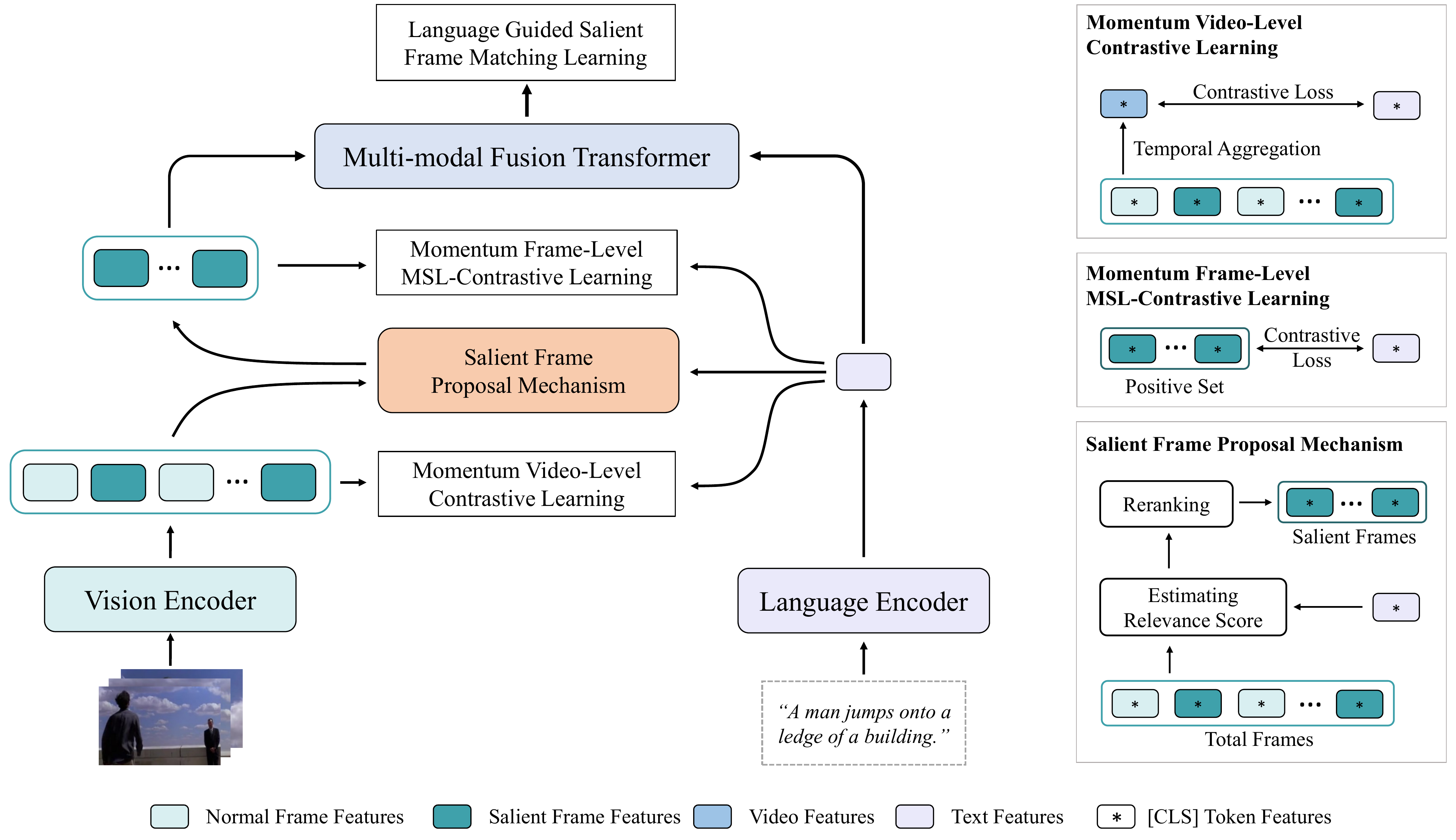}
    \hspace{3.1in} (a) \hspace{2.5in} (b) \hspace{-1.4in}
    \vspace{-0.05in}
    \caption{
    (a) A schematic illustration of the proposed LGDN framework. (b) Details of each component of our LGDN. * denotes that calculation is performed only at the [CLS] token level.
    }
    \label{fig:architecture}
    \vspace{-0.05in}
\end{figure*}

Figure~\ref{fig:architecture} gives a brief overview of our LGDN framework for video-language modeling, which is composed of four main components: 1) language and vision representation extractors; 2) momentum video-level contrastive learning; 3) momentum frame-level MSL-contrastive learning, and 4) language-guided salient frame matching. In the following, we will describe each component in detail.

\subsection{Feature Representation}
\label{sec:representation}

\paragraph{Vision Representation.} 
Given an input video $V$ as a sequence of frames $\{E_i\}_{i=1}^{N}$, where $N$ is the length of the video, we utilize a 2-D vision Transformer (e.g., ViT) as our vision backbone to extract frame-level features $\mathbf{E} = \{\mathbf{E}_1, \mathbf{E}_2, ..., \mathbf{E}_N\}$. Each frame $E_i$ of video $V$ can be represented as $\mathbf{E}_i = [\mathbf{e}_{cls}; \mathbf{e}_1; ...; \mathbf{e}_{k_v-1}] \in R^{k_v\times D_v}$, where $\mathbf{e}_{cls}$ denotes the [CLS] token, $k_v$ denotes the patch sequence length, and $D_v$ denotes the dimension of the patch embeddings. We utilize a fully-connected layer to project the [CLS] token into the frame embedding $\mathbf{f}_i^e$. We then deploy a temporal module $T$ (e.g., a Transformer layer) to aggregate the frame embeddings to obtain the final video embedding:
\begin{equation}
    \mathbf{f}^v =T([\mathbf{f}^e_1, \mathbf{f}^e_2, ..., \mathbf{f}^e_N]) = f^v(V),
\label{attn}
\end{equation}
where $f^v$ denotes the entire vision (video) encoder.

\paragraph{Language Representation.} 
Given an input text $L$, we utilize BERT-Base as our language backbone to extract text feature $\mathbf{L}$, which can be represented as $\mathbf{L} = [\mathbf{l}_{cls}; \mathbf{l}_1; ...;\mathbf{l}_{k_l-1}] \in R^{k_l\times D_l}$, where $\mathbf{l}_{cls}$ is the [CLS] token, $k_l$ is the token sequence length, and $D_l$ is the dimension of the token embeddings. We deploy a fully-connected layer to project the [CLS] token into the text embedding $\mathbf{f}^l = f^l(L)$, where $f^l$ is the language encoder.
 
\subsection{Momentum Video-Level Contrastive Learning (MVCL) Module}
\label{sec:MVCL}

Note that our LGDN is designed to filter out the unmatched/redundant frames for better token-level alignment, without leveraging the temporal information of the videos explicitly. Therefore, we firstly introduce a \textbf{M}omentum \textbf{V}ideo-Level \textbf{C}ontrastive \textbf{L}earning (MVCL) module to address this problem.

The MVCL module utilizes a temporal module (e.g., Transformer block) to aggregate the frame embeddings to obtain the video embedding. Contrastive learning is then applied for holistic video-text alignment. However, video data takes up large GPU memory and the mini-batch size tends to be small with strict resource, which brings harm to contrastive learning. Inspired by MoCo~\cite{he2020moco}, we introduce the momentum mechanism to maintain massive negative samples in memory bank for contrastive learning. Concretely, We firstly maintain video memory bank $\mathcal{M}^v = \{\mathbf{\hat{q}}^v_j\}_{j=1}^{N_m}$ and text memory bank $\mathcal{M}^l = \{\mathbf{\hat{q}}^l_j\}_{j=1}^{N_m}$ to store video/text features, where $N_m$ denotes the memory bank size and $\mathbf{\hat{q}}^v_j$/ $\mathbf{\hat{q}}^l_j$ denotes the $j$-th stored video/text feature vector.
Let $f^v$ (with parameters $\theta^v$) and $\hat{f}^v$ (with parameters $\hat{\theta}^v$) denote vision encoder and vision momentum encoder, respectively. Similarly, let $f^l$ (with parameters $\theta^l$) and $\hat{f}^l$ (with parameters $\hat{\theta}^l$) denote language encoder and language momentum encoder, respectively.
The parameters of momentum encoders are updated by:
\begin{align}
\hat{\theta}^v & = m \cdot \hat{\theta}^v + (1 - m) \cdot \theta^v,~~~~ 
\hat{\theta}^l  = m \cdot \hat{\theta}^l + (1 - m) \cdot \theta^l,
\end{align}
where $m$ is the momentum coefficient hyper-parameter.

The loss function is thus constructed as follow: for each video $V_i$ in mini-batch $\mathcal{B}$, we define the video-to-text contrastive loss between its paired text $L_i$ and all negative samples in the text memory bank $\mathcal{M}^l$, resulting in an InfoNCE loss (with $\tau$ being the temperature hyper-parameter):
\begin{equation}
\mathcal{L}_\text{V2T} \hspace{-1pt}=\hspace{-1pt} -\frac{1}{|\mathcal{B}|} \hspace{-3pt}\sum_{(V_i, L_i) \in \mathcal{B}} \hspace{-7pt}\log \frac{\exp(\cos(\mathbf{f}^v_i, \mathbf{\hat{f}}^l_i) / \tau)}{\exp(\cos(\mathbf{f}^v_i, \mathbf{\hat{f}}^l_i) / \tau) \hspace{-2pt}+\hspace{-2pt} \sum_{\mathbf{\hat{q}}^l_j \in \mathcal{M}^l} \exp(\cos(\mathbf{f}^v_i, \mathbf{\hat{q}}^l_j) / \tau)},
\label{eq:cl_v2t}
\end{equation}
where $\mathbf{\hat{f}}^l_i = \hat{f}^v(l_i)$, and the similarity of two features is measured by the cosine similarity. Similarly, given each text description $L_i$ in mini-batch $\mathcal{B}$, we define the text-to-video contrastive loss as:
\begin{equation}
\mathcal{L}_\text{T2V} \hspace{-1pt}=\hspace{-1pt} -\frac{1}{|\mathcal{B}|} \hspace{-3pt}\sum_{(V_i, L_i) \in \mathcal{B}} \hspace{-7pt}\log \frac{\exp(\cos(\mathbf{f}^l_i, \mathbf{\hat{f}}^v_i) / \tau)}{\exp(\cos(\mathbf{f}^l_i, \mathbf{\hat{f}}^v_i) / \tau) \hspace{-2pt}+\hspace{-2pt} \sum_{\mathbf{\hat{q}}^v_j \in \mathcal{M}^v} \exp(\cos(\mathbf{f}^l_i, \mathbf{\hat{q}}^v_j) / \tau)},
\label{eq:cl_t2v}
\end{equation}
where $\mathbf{\hat{f}}^v_i = \hat{f}^v(V_i)$. 
Finally, the objective function for MVCL is defined as follows:
\begin{equation}
\mathcal{L}_{\text{MVCL}} = \mathcal{L}_{\text{V2T}} + \mathcal{L}_{\text{T2V}}.
\end{equation}

\subsection{Salient Frame Proposal (SFP) Mechanism}

As shown in Figure~\ref{fig:teaser}, video-text data inevitably contains misaligned frame-text pairs. Although an attention mechanism has been applied in Eq.~(\ref{attn}), the irrelevant and noisy information would still mislead the cross-modal alignment in our model. To alleviate this problem, we thus propose a \textbf{S}alient \textbf{F}rame \textbf{P}roposal (SFP) mechanism for video-language modeling. 

The core idea of our SFP mechanism is to dynamically filter out misaligned or redundant frames and maintain only a few important frames to represent the video well, which are called as salient frames. Formally, for each video-text pair, we first identify the relevance score $R(j|i)$ between the text $L_i$ and the $j$-th frame $E_{i,j}$ of the video $V_i$. Further, we perform language-guided denoising to retain only top-$N_{salient}$ salient frames by filtering out the unmatched/redundant frames from each video.

\begin{table}[t]
    \centering
    \caption{Four strategies for estimating relevance scores.}
    \vspace{-0.08in}
    \scalebox{0.82}{
    \tabcolsep8pt
    \begin{tabular}{cccc}
    \toprule
    SimDot & 
    Momentum & 
    CrossMom & 
    Collaborative\\
    \midrule
    $R(j|i) = \mathbf{{f}}^e_{i, j} \cdot \mathbf{{f}}^l_i$ & 
    \text{$R(j|i) = \mathbf{{f}}^e_{i, j} \cdot \mathbf{f}^l_i + \mathbf{\hat{f}}^e_{i, j} \cdot \mathbf{\hat{f}}^l_i$} 
    &
    \text{$R(j|i) = \mathbf{\hat{f}}^e_{i, j} \cdot \mathbf{{f}}^l_i + \mathbf{{f}}^e_{i, j} \cdot \mathbf{\hat{f}}^l_i$}
    &
    \text{$R(j|i) = (\mathbf{{f}}^e_{i, j} + \mathbf{\hat{f}}^e_{i, j}) \cdot(\mathbf{{f}}^l_i + \mathbf{\hat{f}}^l_i)$}\\
    \bottomrule
    \end{tabular}}
    \label{tab:strategies}
    \vspace{-0.05in}
\end{table}

Since only video-level annotations are provided, we need to estimate the relevance scores $R$ automatically. As shown in Table~\ref{tab:strategies}, we introduce four strategies for estimating relevance scores.
\textbf{(1) SimDot prediction} relies on the output of two separate encoders (i.e., frame encoder $f^e$ and language encoder $f^l$) to model the relevance score $R(j|i)$ by computing the dot product of the frame embedding $\mathbf{{f}}^e_{i, j} = f^e(E_{i,j}) $ and the text embedding $\mathbf{{f}}^l_i = f^l(L_i)$. However, since video-text data is noisy, only utilizing single-modality encoders may result in incorrect salient frames.
\textbf{(2) Momentum prediction} improves SimDot prediction by introducing the supervision of momentum encoders (i.e., momentum frame encoder $\hat{f}^e$ and momentum language encoder $\hat{f}^l$), where the momentum frame embedding $\mathbf{\hat{f}}^e_{i, j} = \hat{f}^e(E_{i,j}) $ and the momentum text embedding $\mathbf{\hat{f}}^l_i = \hat{f}^l(L_i)$. 
\textbf{(3) CrossMom prediction} considers the frame-text alignment that is directly built on the interaction between one modality encoder and another modality's momentum encoder. 
\textbf{(4) Collaborative  prediction} combines Momentum prediction and CrossMom prediction for better performance.

Although the frame embeddings and text embeddings can be (roughly) aligned through applying video-text contrastive learning in Sec.~\ref{sec:MVCL}, it is vital to precisely establish frame-text alignment for proposing/selecting salient frames. To this end, we introduce the MFCL module below.

\subsection{Momentum Frame-Level MSL-Contrastive Learning (MFCL) Module}
\label{sec:MFCL}

To dynamically filter out the unmatched/redundant frames, we propose to adopt frame-level contrastive learning to directly measure the relevance scores $R$ between video frames and paired text. However, video data often contains misaligned frame-text pairs. Simply applying standard NCE-based contrastive learning would force the misaligned frame-text pairs to be pulled closer, which inevitably has negative effect on learning high-quality frame-text representation. Inspired by MIL-NCE~\cite{miech2020end}, we thus propose a \textbf{M}omentum \textbf{F}rame-Level Multiple Salient-instance Learning (MSL) \textbf{C}ontrastive \textbf{L}earning (MFCL) module to assist in alleviating the noise problem. The core idea is to use the salient frames filtered by the SFP Mechanism in each video to form a set of positive candidate pairs, instead of considering each positive pair independently. In this work, we suppose that MFCL and SFP have mutual interdependence so that they can bring boost to each other during training. 

Similar to MVCL, we additionally maintain a frame-level memory bank $\mathcal{M}^e = \{\mathbf{\hat{q}}^e_{j'}\}_{{j'}=1}^{N_m * N}$ to store frame features, where $N_m$ is the memory bank size, $N$ is the number of sampled frames per video, and $\mathbf{\hat{q}}^e_{j'}$ is a stored frame feature vector. 

Given each text description $L_i$ in mini-batch $\mathcal{B}$, we select salient frames filtered by the SFP Mechanism in the paired video $V_i$ to form a set of positive candidate (frame-text) pairs $S_i$ and all frame samples in $\mathcal{M}^e$ to form the negative ones. We then define the text-to-frame contrastive loss as:
\begin{equation}
\mathcal{L}_\text{T2E} \hspace{-1pt}=\hspace{-1pt} -\frac{1}{|\mathcal{B}|} \hspace{-3pt}\sum_{(S_i, L_i) \in \mathcal{B}} \hspace{-9pt}\log \frac{\sum_{\mathbf{\hat{f}}^e_{ij} \in \mathbf{\hat{f}}^s_i} \exp(\cos(\mathbf{f}^l_i, \mathbf{\hat{f}}^e_{ij}) / \tau)}{\sum_{\mathbf{\hat{f}}^e_{ij} \in \mathbf{\hat{f}}^s_i} \exp(\cos(\mathbf{f}^l_i, \mathbf{\hat{f}}^e_{ij}) / \tau) \hspace{-2pt}+\hspace{-2pt} \sum_{\mathbf{\hat{q}}^e_{j'} \in \mathcal{M}^e} \exp(\cos(\mathbf{f}^l_{i}, \mathbf{\hat{q}}^e_{j'}) / \tau)},
\end{equation}
where $S_i = \{E_{i,j}\}_{j=1}^N$ is the positive frame set of the video $V_i$, $N$ is the frame sequence length of the video, $\mathbf{f}^l_i = f^l(L_i)$, and $\mathbf{\hat{f}}^s_i = \{\mathbf{\hat{f}}^e_{ij}\}_{j=1}^N= \{\hat{f}^e(E_{i,j})\}_{j=1}^N$.

Similarly, given each positive frame set $S_i$, we define the frame-to-text contrastive loss as:
\begin{equation}
\mathcal{L}_\text{E2T} \hspace{-1pt}=\hspace{-1pt} -\frac{1}{|\mathcal{B}|} \hspace{-3pt}\sum_{(S_i, L_i) \in \mathcal{B}} \hspace{-10pt}\log \frac{\sum_{\mathbf{f}^e_{ij} \in \mathbf{f}^s_i} \exp(\cos(\mathbf{f}^e_{ij}, \mathbf{\hat{f}}^l_i) / \tau)}{\sum_{\mathbf{f}^e_{ij} \in \mathbf{f}^s_i} \exp(\cos(\mathbf{f}^e_{ij}, \mathbf{\hat{f}}^l_i) / \tau) \hspace{-2pt}+\hspace{-2pt}  \sum_{\mathbf{f}^e_{ij} \in \mathbf{f}^s_i} \sum_{\mathbf{\hat{q}}^l_{j'} \in \mathcal{M}^l} \exp(\cos(\mathbf{f}^e_{ij}, \mathbf{\hat{q}}^l_{j'}) / \tau)},
\label{eq:cl_e2t}
\end{equation}
where $\mathbf{\hat{f}}^l_i = \hat{f}^l(L_i)$ and $\mathbf{f}^s_i = \{\mathbf{f}^e_{ij}\}_{j=1}^N = \{f^e(E_{i,j})\}_{j=1}^N$ (text memory bank $\mathcal{M}^l = \{\mathbf{\hat{q}}^l_{j'}\}_{{j'}=1}^{N_m}$ is defined in Sec.~\ref{sec:MVCL}). As a result, by combining the text-to-frame and frame-to-text contrastive losses, the objective function for MFCL is given by:
\begin{equation}
\mathcal{L}_{\text{MFCL}} = \mathcal{L}_{\text{E2T}} + \mathcal{L}_{\text{T2E}}.
\end{equation}

\subsection{Language-Guided Salient Frame Matching (LSFM) Module}
\label{sec:LSFM}

After obtaining language-guided salient frames, we utilize a multi-modal cross-attention fusion Transformer (see Figure~\ref{fig:architecture}) to capture token-level semantic alignment between visual patches and words for better performance (see the design details of this Transformer in the supp. material). 

Further, we take the [CLS] token embedding outputted by the multi-modal fusion Transformer as the joint representation of a frame-text pair ($V_i, L_i$), and deploy a fully-connected layer to predict the matched probability, which is similar to the sentence pair classification task in BERT's pre-training phase. The matching loss is defined as:
\begin{equation}
\mathcal{L}_{\text{LSFM}} = - \mathbb{E}_{(\mathbf{E}_{i, j}, \mathbf{L}_i) \sim  \mathcal{D}_{salient}} \log P(y_{i,j}|\mathbf{E}_{i, j}, \mathbf{L}_i),
\end{equation}
where $\mathbf{E}_{i, j}$ denotes $j$-th frame feature of video $V_i$, $\mathbf{L}_i$ denotes text feature, $\mathcal{D}_{salient}$ is the set of salient frame-text pairs obtained by applying the SFP mechanism to the mini-batch, and $y_{i,j}$ is the ground-truth matching label (0 or 1) of the frame-text pair $(\mathbf{E}_{i, j},\mathbf{L}_i)$. During inference, we use a mean pooling layer to aggregate all salient frame scores as the video-level prediction score. 

Finally, by combining all the proposed modules for video-language modeling at three levels, we train our LGDN model via minimizing the total objective function:
\begin{equation}
\mathcal{L}_{\text{LGDN}} = \mathcal{L}_{\text{MVCL}} + \mathcal{L}_{\text{MFCL}} + \mathcal{L}_{\text{LSFM}}.
\label{eq:total}
\end{equation}

\vspace{-0.3cm}
\section{Experiments}
\vspace{-0.2cm}

\subsection{Datasets and Settings}
\label{exp:setup}
\vspace{-0.1cm}

\noindent\textbf{Pre-Training Datasets.}~~
Due to the restricted computing resources, we follow COTS~\cite{lu2022cots} to pre-train our LGDN on the pure image-text datasets. Our pre-training datasets consists of Conceptual Captions~\cite{sharma2018conceptual}, SBU~\cite{ordonez2011im2text}, VG~\cite{krishna2017visual} and MSCOCO~\cite{lin2014microsoft}, which contains 5.2 million image-text pairs. We additionally apply CC12M~\cite{changpinyo2021conceptual} (about 2 million URLs are now invalid) for better performance, which accumulates 15.2 million image-text pairs in total.

\noindent\textbf{Downstream Datasets.}~~We evaluate our proposed LGDN on four public video-text retrieval datasets: MSR-VTT~\cite{xu2016msr}, MSVD~\cite{chen2011collecting}, DiDeMo~\cite{anne2017localizing}, and VATEX~\cite{wang2021vatex}. 
To further demonstrate the general applicability of our LGDN, we also carry out experiments on a public video-question answering dataset: MSRVTT-QA~\cite{XuZX0Z0Z17}. We present the details of these downstream datasets as well as the evaluation metrics for downstream tasks in the supp. material.

\noindent\textbf{Implementation Details.}~~Following previous work~\cite{lei2021less}, we sample $N = 16$ frames per video: each video is equally split into 16 segments and one frame is randomly sampled from each segment. We empirically set the initial learning rate to 1e-5 and adopt AdamW~\cite{LoshchilovH19} with a weight decay of 0.02 for 5 epochs. 
In the warm-up stage (first epoch), the model is trained to optimize Eq.~(\ref{eq:total}) without applying SFP mechanism. 
We also set the other hyper-parameters uniformly as: salient frame numbers $N_{salient} =2$, mini-batch size $|\mathcal{B}| = 24$, momentum hyper-parameter $m = 0.99$, temperature $\tau = 0.07$, and queue size $N_m = 9,600$. We adopt pre-trained BERT-Base as language encoder and ViT-Base~\cite{alexey2021vit} as vision encoder. More details are given in the supp. material.

\begin{figure}[t!]
    \centering
    \includegraphics[width=0.98\linewidth]{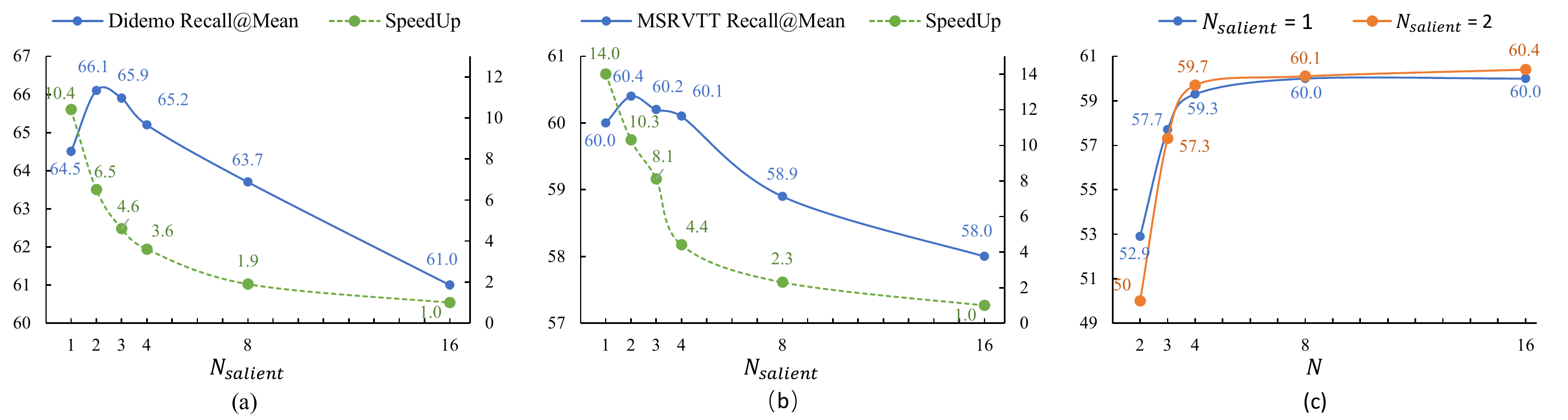}
    \vspace{-0.1in}
    \caption{
    (a-b) Effect of value change of $N_{salient}$. The speedup (green line) is computed w.r.t. the slowest case $N_{salient}=16$ (w/o SFP). We find that utilizing 2--4 salient frames is much more effective while enjoying faster speed. 
    (c) Effect of value change of $N$ on the MSR-VTT 1k-A test set.
    }
    \label{fig:all_salient_frame}
    \vspace{-0.1in}
\end{figure}

\begin{table*}[ht]
    \centering
     \caption{Ablation study for our full LGDN model. Retrieval results are reported on the MSR-VTT 1k-A test set. Local: only token-level alignment during inference. Global: only global alignment during inference. Ensemble: ensemble of global and local (token-level) alignment.  }
    \scalebox{0.78}{
    \tabcolsep3.8pt
    \begin{tabular}{llccccccccc}
    \toprule
    \multirow{2}{*}{Method}& \multirow{2}{*}{Inference}  & \multicolumn{4}{c}{Text-to-Video Retrieval} & \multicolumn{4}{c}{Video-to-Text Retrieval} & \multirow{2}{*}{R@SUM}  \\
    & & R@1 & R@5 & R@10 & MdR & R@1 & R@5 & R@10 & MdR \\
    \midrule 
    $\mathcal{L}_{\text{LSFM}}$ (w/o SFP) & Local &  31.4 & 59.8 & 70.3 & 4.0 & 34.9 & 61.9 & 72.9 & 3.0 & 331.2\\
    $\mathcal{L}_{\text{LSFM}}$ (w/o SFP) $+ \mathcal{L}_{\text{MFCL}}$ (w/o MSL)&  Local & 32.2 & 58.6 & 70.0 & 3.0 & 34.7 & 61.7 & 73.3 & 3.0 & 330.5 \\
    $\mathcal{L}_{\text{LSFM}}$ (w/o SFP) $+ \mathcal{L}_{\text{MFCL}}$& Local & 33.2 & 60.2 & 71.0 & 3.0 & 34.7 & 62.5 & 73.6 & 3.0 & 335.2 \\
    $\mathcal{L}_{\text{LSFM}}$ (w/o SFP) $+ \mathcal{L}_{\text{MFCL}}+ \mathcal{L}_{\text{MVCL}}$& Local & 33.0 & 60.4 & 71.2 & 3.0 & 35.6 & 62.2 & 73.7 & 3.0 & 336.1  \\
    $\mathcal{L}_{\text{LSFM}}$  $+ \mathcal{L}_{\text{MFCL}}+ \mathcal{L}_{\text{MVCL}}$& Local & 35.3 & 65.0 & 75.3 & 3.0 & 36.3 & 65.0 & 76.0 & 3.0 & 352.9 \\
    \midrule
    $\mathcal{L}_{\text{LSFM}}$  $+ \mathcal{L}_{\text{MFCL}}+ \mathcal{L}_{\text{MVCL}}$& Global & 32.5 & 60.4 & 71.7 & 3.0 & 32.1 & 61.8 & 72.2 & 3.0 & 330.7 \\
    $\mathcal{L}_{\text{LSFM}}$  $+ \mathcal{L}_{\text{MFCL}}+ \mathcal{L}_{\text{MVCL}}$ & Ensemble &  \textbf{38.9} & \textbf{65.7}  & \textbf{76.5} & \textbf{2.0} & \textbf{37.9} & \textbf{65.4}  & \textbf{76.0} & \textbf{2.0} & \bf360.4\\
    \bottomrule
    \end{tabular}}
    \label{tab:main_ablation}
    \vspace{-0.1in}
\end{table*}

\noindent\textbf{Evaluation Metrics.}~~We adopt two widely-used metrics in cross-modal retrieval: Recall at K (R@K, K$=1,5,10$), and Median Rank (MdR) / Mean Rank (MnR). R@K means the percentage of correct matching in the K nearest points, and MdR / MnR measures the median / mean rank of target items in the retrieved ranking list. We also report two additional metrics named `R@Sum' and `R@Mean' in our ablation study, which sums/averages all recall metrics for overall evaluation. Following ClipBERT~\cite{lei2021less}, we also report accuracy (Acc) in video-question answering task.

\subsection{Ablation Study}

In this subsection, we conduct comprehensive ablation study to investigate the contributions of different components of our full model. If not specifically indicated, we set $N = 16$ for global alignment and $N_{salient} = 2$ for token-level alignment as the default setting.

\noindent\textbf{Effect of Value Change of $N_{salient}$ and $N$.} 
A common perspective for video/video-language understanding is that more frames per video bring better performance. We thus conduct experiments on the frame number used for token-level alignment in Figure~\ref{fig:all_salient_frame}(a-b). 
We sample $N = 16$ frames from each video and evaluate different variants that use $N_{salient} \in \{1, 2, 3, 4, 8, 16\}$ frames. Note that when $N_{salient} = 16$, sampling by our SFP degrades to w/o SFP. It can be observed that utilizing only $N_{salient} = \{2, 3, 4\}$ salient frames filtered by our SFP significantly outperforms utilizing all 16 extracted frames meanwhile enjoying the faster speed (see the green lines). This suggests that our SFP mechanism not only selects correct salient frames but also alleviates the noise problem.
To investigate the influence of value change of $N$ on our LGDN, we evenly sample $N \in \{2, 3, 4, 8\}$ frames per video and freeze $N_{salient} = \{1, 2\}$  salient frames. The results in Figure~\ref{fig:all_salient_frame}(c) indicate that more extracted frames per video are beneficial to the token-level alignment in our LGDN model, as it provides larger candidate set for selecting salient frames. Meanwhile, when $N$ becomes larger (> 4), the performance tends to converge, further demonstrating the redundancy in the videos.

\noindent\textbf{Contributions of Each Components.}
We further demonstrate the contributions of the three objective functions as well as the salient frame proposal (SFP) mechanism used in our full LGDN model in Table~\ref{tab:main_ablation}. We start with the objective function $\mathcal{L}_{\text{LSFM}}$ (w/o SFP), which means only applying matching loss in token-level alignment without using the SFP mechanism. It can be observed that: (1) $\mathcal{L}_{\text{MFCL}}$ (and $\mathcal{L}_{\text{MVCL}}$) combined with $\mathcal{L}_{\text{LSFM}}$ (w/o SFP) can bring improvements, suggesting that global alignment is beneficial to token-level alignment (during the training stage). (2) Simply applying the frame-level alignment may cause negative effect while combing with our MSL design brings better results. This demonstrates that our design of $\mathcal{L}_{\text{MFCL}}$ does help alleviate the noise problem. (3) When the SFP mechanism is added (see $\mathcal{L}_{\text{LSFM}}$ (w/o SFP) $+ \mathcal{L}_{\text{MFCL}}+ \mathcal{L}_{\text{MVCL}}$ vs. $\mathcal{L}_{\text{LSFM}}$  $+ \mathcal{L}_{\text{MFCL}}+ \mathcal{L}_{\text{MVCL}}$), the performance is significantly improved, which clearly shows the effectiveness of our proposed SFP mechanism. (4) For the same trained full LGDN model, combining the global and token-level alignment during inference can bring further improvements. Note that our full LGDN still achieves the state-of-the-art on MSR-VTT even without considering global alignment during inference.

\begin{table}[t]
\centering
\caption{Comparison to the state-of-the-arts for video-text retrieval on MSR-VTT. Extra Expert: methods utilized expert features (e.g., object, motion and OCR features). \# PT Pairs: the number of pre-training pairs. $^\dagger$ denotes that our LGDN is additionally pre-trained with CC12M~\cite{changpinyo2021conceptual}. }
\vspace{-0.08in}
\scalebox{0.78}{
\tabcolsep8pt
\begin{tabular}{lcrcccccccc}
\toprule
\multirow{2}{*}{Method}  & Extra  & \multirow{2}{*}{\#PT Pairs} & \multicolumn{4}{c}{Text-to-Video Retrieval} & \multicolumn{4}{c}{Video-to-Text Retrieval}  \\
& Expert & & R@1 & R@5 & R@10 & MdR & R@1 & R@5 & R@10 & MdR  \\
\midrule
\textbf{Full Split:} & \\
HGR~\cite{chen2020fine} &  & - & 9.2 & 26.2 & 36.5 & 24.0 & 15.0 & 36.7 & 48.8 & 11.0 \\
CE~\cite{liu2019use} & $\checkmark$ & - & 10.0 & 29.0 & 41.2 & 16.0 & 15.6 & 40.9 & 55.2 & 8.3 \\ 
CMGSD~\cite{he2021improving} & \checkmark & $>$100M  & 11.3 & 32.0 & 44.1 & 14.2  & 17.2 & 43.6 & 57.2 & 7.6 \\ 
T2VLAD~\cite{wang2021t2vlad} & \checkmark & $>$100M  & 12.7 & 34.8 & 47.1 & 12.0  & 20.7 & 48.9 & 62.1 & 6.0 \\ 

\textbf{LGDN~(ours)} &  & 5.2M & \textbf{22.9} & \textbf{46.0}  & \textbf{56.8} & \textbf{7.0} & \textbf{41.8} & \textbf{65.2}  & \textbf{74.6} & \textbf{2.0}\\
\textbf{LGDN$^\dagger$~(ours)} &  &  15.2M & \textbf{27.5} & \textbf{51.7}  & \textbf{61.9} & \textbf{5.0} & \textbf{50.2} & \textbf{73.9}  & \textbf{82.3} & \textbf{1.0} \\
\midrule
\textbf{7k-1k Split:} & \\

HERO~\cite{li2020hero} &  &  $>$100M & 16.8 & 43.4 & 57.7 & - & - & - & - & - \\

UniVL~\cite{luo2020univilm} & $\checkmark$ & $>$100M & 21.2 & 49.6 & 63.1 & 6.0 & - & - & - & - \\ 
ClipBERT~\cite{lei2021less} &  & 5.6M & 22.0 & 46.8 & 59.9 & 6.0  & - & - & - & - \\ 
TACo~\cite{yang2021taco} & $\checkmark$ &  $>$100M & 24.8 & 52.1 & 64.5 & 5.0 & - & - & - & -\\
\textbf{LGDN~(ours)} &   & 5.2M & \textbf{34.3} & \textbf{62.5}  & \textbf{72.2} & \textbf{3.0} & \textbf{34.7} & \textbf{60.8}  & \textbf{70.4} & \textbf{3.0}\\
\textbf{LGDN$^\dagger$~(ours)} &  & 15.2M & \textbf{39.8} & \textbf{65.2}  & \textbf{77.0} & \textbf{2.0} & \textbf{39.2} & \textbf{66.4}  & \textbf{76.1} & \textbf{3.0}\\
\midrule
\textbf{1k-A Split:} & \\

MMT~\cite{gabeur2020multi} &\checkmark  &   $>$100M & 26.6 & 57.1 & 69.6 & 4.0 & 27.0 & 57.5 & 69.7 & 3.7\\
Support Set~\cite{patrick2020support} & \checkmark  & $>$100M & 30.1 & 58.5 &69.3 & 3.0  & 28.5 & 58.6 & 71.6 & 3.0\\

TACo~\cite{yang2021taco} &\checkmark& $>$100M & 28.4 & 57.8 & 71.2 & 4.0 \\
Frozen in Time~\cite{bain2021frozen} &  &  5.5M & 31.0 & 59.5 & 70.5 & 3.0 & - & - & - & -\\ 
\textbf{LGDN~(ours)} &   & 5.2M & \textbf{38.9} & \textbf{65.7}  & \textbf{76.5} & \textbf{2.0} & \textbf{37.9} & \textbf{65.4}  & \textbf{76.0} & \textbf{2.0}\\
\textbf{LGDN$^\dagger$~(ours)} &   & 15.2M  & \textbf{43.7} & \textbf{71.4}  & \textbf{80.4} & \textbf{2.0}& \textbf{42.6} & \textbf{71.6}  & \textbf{80.6} & \textbf{2.0}\\

\bottomrule
\end{tabular}}
\label{tab:msr-vtt-sota}
\vspace{-0.1in}
\end{table}

\begin{figure}
\centering
\begin{minipage}[t]{0.48\textwidth}
   \centering
        \makeatletter\def\@captype{table}\makeatother
        \caption{Results on the VATEX test set.}
        \vspace{0.05in}
        \scalebox{0.8}{
        \tabcolsep6pt
        \begin{tabular}{@{}lcccc@{}}
        \toprule
        {Method}  & {R@1 }  & {R@5 }  & {R@10 } & {MdR }  \\
        \midrule
        VSE~\cite{kiros2014unifying} & 28.0 & 64.3 & 76.9 & 3.0 \\
        VSE++~\cite{faghri2017vse++} & 33.7 & 70.1 & 81.0 & 2.0 \\
        Dual~\cite{mithun2018learning} & 31.1 & 67.4 & 78.9 & 3.0 \\
        HGR~\cite{chen2020fine} & 35.1 & 73.5 & 83.5 & 2.0 \\
        Support Set~\cite{patrick2020support} & 45.9 & 82.4 & 90.4 & \textbf{1.0} \\
         \midrule
        \textbf{LGDN~(ours)} & \textbf{57.1} & \textbf{87.5} & \textbf{93.6} & \textbf{1.0}  \\ 
        \textbf{LGDN$^\dagger$~(ours)} & \textbf{61.0} & \textbf{90.2} & \textbf{95.1} & \textbf{1.0}  \\
        \bottomrule
        \end{tabular}}
        \label{tab:vatex}
   \end{minipage}
   \vspace{0.1in}
   \hspace{0.08in}
   \begin{minipage}[t]{0.46\textwidth}
    \centering
     \makeatletter\def\@captype{table}\makeatother
        \caption{Results on the MSVD test set.}
        \vspace{0.05in}
        \scalebox{0.8}{
        \tabcolsep4pt
        \begin{tabular}{@{}lcccc@{}}
        \toprule
        {Method}  & {R@1 }  & {R@5 }  & {R@10 } & {MdR } \\
        \midrule
        VSE++~\cite{faghri2017vse++} & 15.4 & 39.6 & 53.0 & 9.0 \\
        Multi. Cues~\cite{mithun2018learning} & 20.3 & 47.8 & 61.1 & 6.0 \\
        CE~\cite{liu2019use} & 19.8 & 49.0 & 63.8 & 6.0 \\
        Support Set~\cite{patrick2020support} & 28.4 & 60.0 & 72.9 & 4.0 \\
        Frozen in Time~\cite{bain2021frozen} & 33.7 & 64.7 & 76.3 & 3.0 \\ 
         \midrule
        \textbf{LGDN~(ours)} & \textbf{39.7} & \textbf{70.2} & \textbf{79.8} & \textbf{2.0}  \\ 
        \textbf{LGDN$^\dagger$~(ours)} & \textbf{43.2} & \textbf{73.3} & \textbf{82.4} & \textbf{2.0}  \\ 
        \bottomrule
        \end{tabular}}
        \label{tab:msvd}
  \end{minipage}
  \vspace{0.1in}
  \begin{minipage}[t]{0.48\textwidth}
  \centering
     \makeatletter\def\@captype{table}\makeatother
        \caption{Results on the DiDeMo test set. $^*$ denotes using temporal labels of captions. }
        \vspace{0.05in}
        \scalebox{0.8}{
        \tabcolsep5pt
        \begin{tabular}{@{}lcccc@{}}
        \toprule
        {Method}  & {R@1 }  & {R@5 }  & {R@10 } & {MdR } \\
        \midrule
        S2VT~\cite{venugopalan2014translating} & 11.9 & 33.6 & - & 13.0 \\
        FSE~\cite{zhang2018cross} & 13.9 & 36.0 & - & 11.0 \\
        CE~\cite{liu2019use} & 16.1 & 41.1 & - & 8.3 \\
        ClipBERT~\cite{lei2021less}$^*$ & 20.4 & 48.0 & 60.8 & 6.0 \\
        Frozen in time~\cite{bain2021frozen}$^*$ & 34.6 & 65.0 & 74.7 & \textbf{2.0} \\
        \midrule
        \textbf{LGDN~(ours)} & \textbf{44.1} & \textbf{71.9} & \textbf{82.3} & \textbf{2.0}  \\ 
        \textbf{LGDN$^\dagger$~(ours)} & \textbf{47.8} & \textbf{76.2} & \textbf{83.3} & \textbf{2.0}  \\ 
        \bottomrule
        \end{tabular}}
        \label{tab:didemo}
  \end{minipage}
  \hspace{0.1in}
  \begin{minipage}[t]{0.46\textwidth}
   \centering
        \makeatletter\def\@captype{table}\makeatother
        \caption{Results on MSRVTT-QA. $^*$ denotes utilizing large-scale VideoQA datasets.
        }
        \vspace{0.05in}
        \scalebox{0.8}{
        \tabcolsep8pt
        \begin{tabular}{@{}lcccc@{}}
        \toprule
        {Method}& \#PT Pairs  & Acc  \\
        \midrule
        Heterogeneous Memory~\cite{FanZZW0H19} & - & 33.0  \\
        HCRN~\cite{LeLV020} & - & 35.6  \\
        SSML~\cite{AmraniBRB21} & 100M & 35.1 \\
        ClipBERT~\cite{lei2021less} & 5.6M & 37.4  \\
        Just Ask$^*$~\cite{YangMSLS21} & 69.0M & 41.5  \\
         \midrule
        \textbf{LGDN~(ours)} & 5.2M & \textbf{42.4} \\ 
        \textbf{LGDN$^\dagger$~(ours)} & 15.2M & \textbf{43.1}\\
        \bottomrule
        \end{tabular}}
        \label{tab:msrvttqa}
   \end{minipage}
   \vspace{-0.2in}
\end{figure}

\vspace{-0.15cm}
\subsection{Comparison to the State-of-the-Arts}
\vspace{-0.15cm}

We first report the text-video retrieval results on MSR-VTT with three data partitions in Table~\ref{tab:msr-vtt-sota}. It can be observed that: 
(1) our LGDN outperforms all previous works by large margins. Particularly, as compared with the most recent model Frozen in Time~\cite{bain2021frozen}, our LGDN achieves an improvement of 7.9\% (38.9\% vs. 31.0\%) for Text-to-Video R@1 on the MSR-VTT 1k-A test set. 
(2) Our LGDN also outperforms methods utilizing extra modalities (e.g., motion and audio) or those pre-trained on extremely-large video data (e.g., HowTo100M). 
(3) When leveraging a much larger pre-training (image-text) dataset, our LGDN (marked with $^\dagger$) achieves significant improvements.

To demonstrate the robustness of our model, we also evaluate it on VATEX, MSVD, and Didemo in Tables~\ref{tab:vatex}--\ref{tab:didemo}, respectively. Due to limited space, only text-to-video retrieval is considered here. For VATEX (Table~\ref{tab:vatex}), our LGDN significantly outperforms the state-of-the-art method Support Set which is trained on an order of magnitude more data. Our LGDN still performs the best on MSVD (Table~\ref{tab:msvd}) and Didemo (Table~\ref{tab:didemo}). 
Particularly, in the Didemo dataset, each description is annotated with localization information, in other words, annotations may only be aligned with the localized moments, thus causing the noise problem as many methods utilize all frames as the input. Recent works exploit temporal labels of captions to alleviate the noise problem and achieve higher performance. However, even without considering this, our LGDN still largely outperforms the most recent method Frozen~\cite{bain2021frozen}, further demonstrating the effectiveness of our LGDN. 

To show the general applicability of our LGDN, we evaluate our LGDN on the VideoQA task in Table~\ref{tab:msrvttqa}. Even without utilizing large-scale video datasets devoted to the VideoQA task, our LGDN outperforms all competitors, validating the effectiveness of our LGDN in VideoQA. In addition, to reveal the critical importance of solving the noise issue for video-language modeling, we directly apply the SFP mechanism to the latest model CLIP4Clip~\cite{luo2021clip4clip} in Table~\ref{tab:clip4clip}. We find that applying the SFP mechanism brings boost to Clip4CLIP. The ensemble mechanism further improves the results, indicating that the proposed SFP mechanism is complementary to the baseline.

\begin{table}[t]
\centering
\caption{
Further evaluation results by directly applying our SFP mechanism to Clip4CLIP~\cite{luo2021clip4clip} (re-implemented) for video-text retrieval on the MSR-VTT 1k-A test set. $^*$ denotes the ensemble results of Clip4CLIP and Clip4CLIP+SFP.}
\vspace{-0.08in}
\scalebox{0.86}{
\tabcolsep7.6pt
\begin{tabular}{lcccccccccc}
\toprule
\multirow{2}{*}{Method} & \multicolumn{5}{c}{Text-to-Video Retrieval} & \multicolumn{5}{c}{Video-to-Text Retrieval}  \\
&  R@1 & R@5 & R@10 & MdR & MnR & R@1 & R@5 & R@10 & MdR & MnR \\
\midrule
Clip4CLIP~\cite{luo2021clip4clip}  & {44.9} & {71.8}  & {81.7} & {2.0} & 14.2  & {46.2} & {73.9}  & {84.3} & {2.0} & 10.8\\
\midrule
\: +\: SFP    & {45.3} & {73.0}  & {83.4} & {2.0} & 13.4 & {47.6} & {75.5}  & {85.3} & {2.0} & 9.6\\
\: +\: SFP$^*$  & \textbf{47.2} & \textbf{73.4}  & \textbf{83.9} & \textbf{2.0}& \textbf{13.0} & \textbf{48.1} & \textbf{76.7}  & \textbf{86.1} & \textbf{2.0} & \textbf{9.3}\\
\bottomrule
\end{tabular}}
\label{tab:clip4clip}
\vspace{-0.0in}
\end{table}

\begin{table}[t!]
\centering
    \caption{Results by applying SFP to different sampling techniques on the MSR-VTT 1kA test set.}
    \label{tab:sampling}
    \vspace{-0.08in}
    \scalebox{0.86}{
    \tabcolsep13pt
        \begin{tabular}{@{}l|ccc|ccc@{}}
        \toprule
        \multirow{2}{*}{Sampling}   & \multicolumn{3}{|c|}{w\textbackslash{}o SFP (R@SUM)}  & \multicolumn{3}{c}{w\textbackslash SFP (R@SUM)} \\
         & 4 frames   & 8 frames   & 16 frames    & 4 frames & 8 frames  & 16 frames \\
         \midrule
        Random Sampling & 164.7 & 169.5 & 172.8 & 168.0    & 174.3    & 179.4    \\
        Dense Uniform          & 166.5     & 171.3     & 174.3     & 173.1    & 179.4    & 180.6    \\
        Sparse Sampling        & 168.0     & 171.9     & 173.7     & 179.1    & 180.3    & 181.1   \\
        \bottomrule
        \end{tabular}}
\end{table}

\begin{table}[t!]
    \centering
    \caption{Ablation study for relevance score estimator. Retrieval results are reported on the MSR-VTT 1k-A test set. Random / SFP: $N_{salient}$ frames are sampled randomly/by our SFP mechanism. \# Frames: $N_{salient}$  / $N$ for local / global alignment.}
    \vspace{-0.08in}
    \scalebox{0.86}{
    \tabcolsep4.8pt
    \begin{tabular}{lccccccccccc}
    \toprule
    \multirow{2}{*}{Method} & \multirow{2}{*}{Strategy}& \multirow{2}{*}{\# Frames}  & \multicolumn{4}{c}{Text-to-Video Retrieval} & \multicolumn{4}{c}{Video-to-Text Retrieval} & \multirow{2}{*}{R@SUM} \\
    & & & R@1 & R@5 & R@10 & MdR & R@1 & R@5 & R@10  & MdR  \\
    \midrule
    All & Random & 16/16 & 35.7 & 63.8 & 74.2 & 3.0 & 35.4 & 63.8 & 74.9 & 3.0 & 347.8 \\
    Random & Random & 2/16 & 34.1 & 62.1 & 73.4 & 3.0 & 34.3 & 61.6 & 74.0 & 3.0 & 339.5\\
    \midrule
    SimDot & SFP & 2/16 & 37.4 & 65.0 & 76.4 & 3.0 & 37.2 & 65.1 & 75.4 & 2.0 & 356.5\\
    Momentum & SFP & 2/16 & 38.1 & \bf65.8 & 76.4 & 2.0 & 37.9 & 65.4 & 75.9 & 2.0 & 359.5\\
    CrossMom & SFP & 2/16 & 38.4 & 65.4 & 76.5 & 2.0 & 37.9 & 65.3 & \bf76.2 & 2.0 &  359.7 \\   
    Collaborative & SFP & 2/16 & \bf38.9 & 65.7 & \bf76.5 & \bf2.0 & \bf37.9 & \bf65.4 & 76.0 & \bf2.0 & \bf360.4\\
    \bottomrule
    \end{tabular}}
    \label{tab:ablation_relevance}
    \vspace{-0.1in}
\end{table}

\subsection{Additional Results}

\noindent\textbf{Applying SFP to Different Frame Sampling Techniques.} 
Note that our SFP mechanism must be combined with a frame sampling technique since we adopt a two-stage sampling strategy in this paper. Thus, we apply our SFP mechanism to three frame sampling techniques: Sparse Sampling, Random Sampling, and Dense Uniform (equally interval sampling). The obtained results on the MSR-VTT 1kA test set are provided in Table~\ref{tab:sampling}. It can be observed that our SFP significantly boosts different sampling strategies, further demonstrating the general applicability of our SFP mechanism.

\noindent\textbf{Expansion of Relevance Score Estimator.}
In Sec.~3.3, we have proposed four relevance score estimators for the LSFM module. To find out which is the best, we present the ablation study results for different relevance score estimators in Table~\ref{tab:ablation_relevance}. We can see a large gap between SFP and random sampling (w/o SFP), directly demonstrating the effectiveness of the proposed SFP mechanism. Meanwhile, both Momentum and CrossMom outperform SimDot, suggesting that introducing momentum encoder is beneficial to relevance score estimation. Collaborative that combines Momentum and CrossMom generally leads to further improvements.

\noindent\textbf{Model Capacity.}
We also provide the detailed comparison to other methods in terms of model capacity and R@SUM (on the MSR-VTT 1kA test set) in Table~\ref{tab:capacity}. It can be clearly seen that: (i) When fusion layers are not used (i.e., only global alignment is adopted), our LGDN (global) outperforms the state-of-the-art method Frozen in Time~\cite{bain2021frozen}, but with much less model parameters. (ii) Our full LGDN performs much better than all the competitors, but its parameter number (215M) is still comparable to that of Frozen in Time (180M) and even significantly smaller than those of the other competitors. These observations suggest that the performance gains obtained by our LGDN is not due to utilizing more model parameters.

\begin{table}[t!]
\centering
    \caption{The model capacity of different recent methods on the MSR-VTT 1kA test set.}
    \label{tab:capacity}
    \vspace{-0.08in}
    \scalebox{0.85}{
    \tabcolsep10pt
    \begin{tabular}{@{}lccccc@{}}
        \toprule
        Methods & Visual Encoder & Lingual Encoder & Fusion Layer & Total & R@SUM \\
         \midrule
        TACo~\cite{yang2021taco} & 155M& 110M & 14M          & 279M  & 157.4 \\
        Support Set~\cite{patrick2020support}    & 136M& 220M & - & 356M  & 157.9 \\
        Frozen in Time~\cite{bain2021frozen} & 114M& 66M  & - & 180M  & 161.0 \\
        LGDN (global)  & 93M & 55M  & - & \bf{148M}  & 164.6 \\
        LGDN (ours)    & 93M & 55M  & 68M          & 215M  & \bf{181.1} \\
        \bottomrule
    \end{tabular}}
    \vspace{-0.05in}
\end{table}

\begin{figure*}[t]
    \centering
    \includegraphics[width=0.98\textwidth]{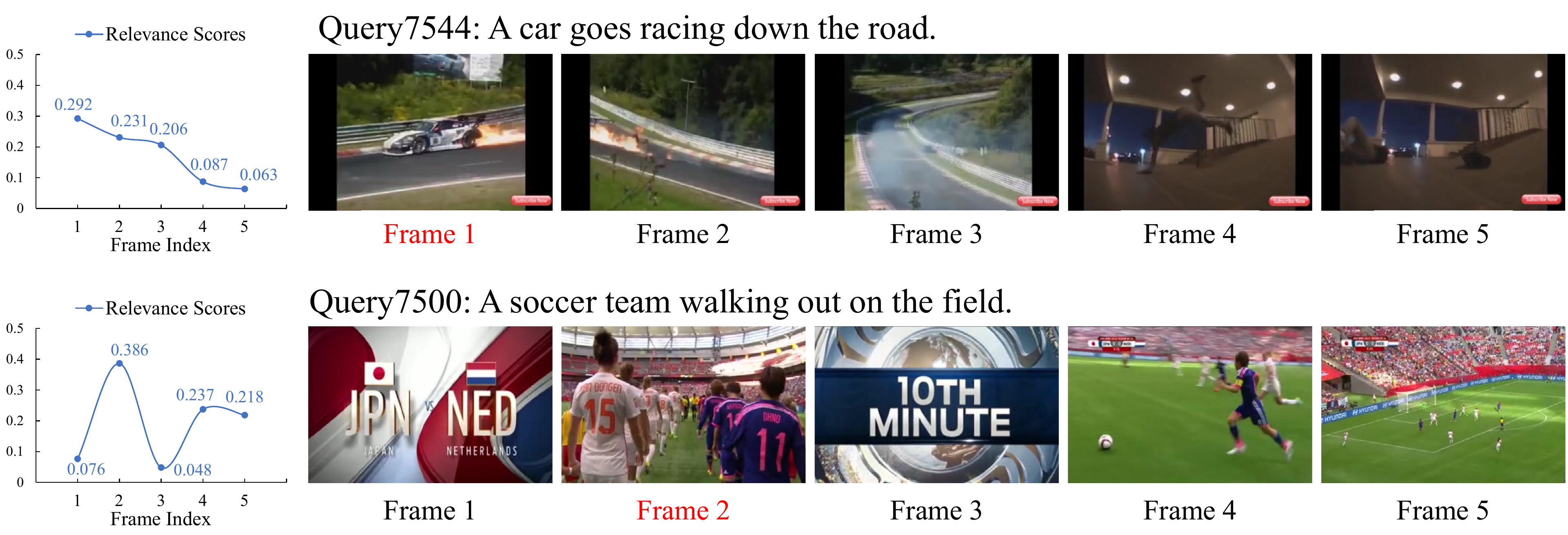}
    \vspace{-0.1in}
    \caption{
    Visualization of our LGDN on the MSR-VTT test set. We uniformly sample 5 frames from each video and the red denotes salient frames selected by our SFP mechanism. We find that our SFP mechanism indeed filters out the unmatched/redundant frames under the language supervision.
    }
    \label{fig:visualize_all}
    \vspace{-0.1in}
\end{figure*}

\subsection{Visualization Results}

We provide visualization of our LGDN in Figure~\ref{fig:visualize_all}.
We uniformly sample 5 frames from each video and provide relevance scores of 5 frames on the left, and the red ones denote salient frames selected by the SFP mechanism. It can be seen that: (1) Although the holistic video is semantically related to the paired text, there still exist noisy frames (e.g., the transition in Frame 1 and Frame 3 of Query7500) and unrelated frames (e.g., in Frame 4 and Frame 5 of Query7544, a man is rolling while the paired text is `a car goes racing down the road'). (2) The relevance scores obtained from the SFP mechanism correctly measure the consistency between each frame and the paired text, which indeed helps our LGDN to precisely filter out noisy information for better video-language modeling.

\section{Conclusion}

In this work, we propose a novel Language-Guided Denoising Network (LGDN) for video-language modeling, which can dynamically filter out the unmatched or redundant frames under the language supervision and thus maintain only 2--4 salient frames per video for cross-modal token-level alignment. Extensive experiments on five public datasets show that our LGDN outperforms the state-of-the-arts by large margins. In the future, we will consider aggregating temporal information on salient frames and apply our approach to more challenging video-language tasks (e.g., video grounding).

\begin{ack}
This work was supported in part by National Natural Science Foundation of China (61976220 and 61832017), Beijing Outstanding Young Scientist Program (BJJWZYJH012019100020098), and the Research Seed Funds of School of Interdisciplinary Studies, Renmin University of China.
\end{ack}

{\small
\bibliographystyle{plain}  
\bibliography{LGDN}

\begin{thebibliography}{10}

\bibitem{AmraniBRB21}
Elad Amrani, Rami Ben{-}Ari, Daniel Rotman, and Alex~M. Bronstein.
\newblock Noise estimation using density estimation for self-supervised
  multimodal learning.
\newblock In {\em AAAI}, pages 6644--6652, 2021.

\bibitem{bain2021frozen}
Max Bain, Arsha Nagrani, G{\"u}l Varol, and Andrew Zisserman.
\newblock Frozen in time: A joint video and image encoder for end-to-end
  retrieval.
\newblock In {\em ICCV}, pages 1728--1738, 2021.

\bibitem{changpinyo2021conceptual}
Soravit Changpinyo, Piyush Sharma, Nan Ding, and Radu Soricut.
\newblock {Conceptual 12M}: Pushing web-scale image-text pre-training to
  recognize long-tail visual concepts.
\newblock In {\em CVPR}, pages 3558--3568, 2021.

\bibitem{chen2011collecting}
David Chen and William~B Dolan.
\newblock Collecting highly parallel data for paraphrase evaluation.
\newblock In {\em ACL}, pages 190--200, 2011.

\bibitem{chen2020fine}
Shizhe Chen, Yida Zhao, Qin Jin, and Qi~Wu.
\newblock Fine-grained video-text retrieval with hierarchical graph reasoning.
\newblock In {\em CVPR}, pages 10635--10644, 2020.

\bibitem{chen2020simple}
Ting Chen, Simon Kornblith, Mohammad Norouzi, and Geoffrey Hinton.
\newblock A simple framework for contrastive learning of visual
  representations.
\newblock In {\em ICML}, pages 1597--1607, 2020.

\bibitem{jacob2019bert}
Jacob Devlin, Ming{-}Wei Chang, Kenton Lee, and Kristina Toutanova.
\newblock {BERT}: Pre-training of deep bidirectional transformers for language
  understanding.
\newblock In {\em NAACL-HLT}, pages 4171--4186, 2019.

\bibitem{dong2018predicting}
Jianfeng Dong, Xirong Li, and Cees~GM Snoek.
\newblock Predicting visual features from text for image and video caption
  retrieval.
\newblock {\em IEEE Transactions on Multimedia}, 20(12):3377--3388, 2018.

\bibitem{alexey2021vit}
Alexey Dosovitskiy, Lucas Beyer, Alexander Kolesnikov, Dirk Weissenborn,
  Xiaohua Zhai, Thomas Unterthiner, Mostafa Dehghani, Matthias Minderer, Georg
  Heigold, Sylvain Gelly, Jakob Uszkoreit, and Neil Houlsby.
\newblock An image is worth 16x16 words: Transformers for image recognition at
  scale.
\newblock In {\em ICLR}, 2021.

\bibitem{faghri2017vse++}
Fartash Faghri, David~J. Fleet, Jamie~Ryan Kiros, and Sanja Fidler.
\newblock {VSE++:} improving visual-semantic embeddings with hard negatives.
\newblock In {\em BMVC}, page~12, 2018.

\bibitem{FanZZW0H19}
Chenyou Fan, Xiaofan Zhang, Shu Zhang, Wensheng Wang, Chi Zhang, and Heng
  Huang.
\newblock Heterogeneous memory enhanced multimodal attention model for video
  question answering.
\newblock In {\em CVPR}, pages 1999--2007, 2019.

\bibitem{gabeur2020multi}
Valentin Gabeur, Chen Sun, Karteek Alahari, and Cordelia Schmid.
\newblock Multi-modal transformer for video retrieval.
\newblock In {\em ECCV}, pages 214--229, 2020.

\bibitem{he2021improving}
Feng He, Qi~Wang, Zhifan Feng, Wenbin Jiang, Yajuan Lu, Yong Zhu, and Xiao Tan.
\newblock Improving video retrieval by adaptive margin.
\newblock In {\em SIGIR}, pages 1359--1368, 2021.

\bibitem{he2020moco}
Kaiming He, Haoqi Fan, Yuxin Wu, Saining Xie, and Ross~B. Girshick.
\newblock Momentum contrast for unsupervised visual representation learning.
\newblock In {\em CVPR}, pages 9726--9735, 2020.

\bibitem{anne2017localizing}
Anne~Lisa Hendricks, Oliver Wang, Eli Shechtman, Josef Sivic, Trevor Darrell,
  and Bryan Russell.
\newblock Localizing moments in video with natural language.
\newblock In {\em ICCV}, pages 5804--5813, 2017.

\bibitem{hermans2017defense}
Alexander Hermans, Lucas Beyer, and Bastian Leibe.
\newblock In defense of the triplet loss for person re-identification.
\newblock {\em arXiv preprint arXiv:1703.07737}, 2017.

\bibitem{huo2021wenlan}
Yuqi Huo, Manli Zhang, Guangzhen Liu, Haoyu Lu, Yizhao Gao, Guoxing Yang,
  Jingyuan Wen, Heng Zhang, Baogui Xu, Weihao Zheng, et~al.
\newblock {WenLan}: Bridging vision and language by large-scale multi-modal
  pre-training.
\newblock {\em arXiv preprint arXiv:2103.06561}, 2021.

\bibitem{jia2021scaling}
Chao Jia, Yinfei Yang, Ye~Xia, Yi-Ting Chen, Zarana Parekh, Hieu Pham, Quoc~V
  Le, Yunhsuan Sung, Zhen Li, and Tom Duerig.
\newblock Scaling up visual and vision-language representation learning with
  noisy text supervision.
\newblock In {\em ICML}, pages 4904--4916, 2021.

\bibitem{jin2021hierarchical}
Weike Jin, Zhou Zhao, Pengcheng Zhang, Jieming Zhu, Xiuqiang He, and Yueting
  Zhuang.
\newblock Hierarchical cross-modal graph consistency learning for video-text
  retrieval.
\newblock In {\em SIGIR}, pages 1114--1124, 2021.

\bibitem{kim2021vilt}
Wonjae Kim, Bokyung Son, and Ildoo Kim.
\newblock {ViLT}: Vision-and-language transformer without convolution or region
  supervision.
\newblock In {\em ICML}, pages 5583--5594, 2021.

\bibitem{kiros2014unifying}
Ryan Kiros, Ruslan Salakhutdinov, and Richard~S. Zemel.
\newblock Unifying visual-semantic embeddings with multimodal neural language
  models.
\newblock {\em arXiv preprint arXiv:1411.2539}, 2014.

\bibitem{krishna2017visual}
Ranjay Krishna, Yuke Zhu, Oliver Groth, Justin Johnson, Kenji Hata, Joshua
  Kravitz, Stephanie Chen, Yannis Kalantidis, Li-Jia Li, David~A Shamma, et~al.
\newblock Visual genome: Connecting language and vision using crowdsourced
  dense image annotations.
\newblock {\em IJCV}, 123(1):32--73, 2017.

\bibitem{LeLV020}
Thao~Minh Le, Vuong Le, Svetha Venkatesh, and Truyen Tran.
\newblock Hierarchical conditional relation networks for video question
  answering.
\newblock In {\em CVPR}, pages 9972--9981, 2020.

\bibitem{lei2021less}
Jie Lei, Linjie Li, Luowei Zhou, Zhe Gan, Tamara~L Berg, Mohit Bansal, and
  Jingjing Liu.
\newblock Less is more: {ClipBERT} for video-and-language learning via sparse
  sampling.
\newblock In {\em CVPR}, pages 7331--7341, 2021.

\bibitem{li2021albef}
Junnan Li, Ramprasaath Selvaraju, Akhilesh Gotmare, Shafiq Joty, Caiming Xiong,
  and Steven Chu~Hong Hoi.
\newblock Align before fuse: Vision and language representation learning with
  momentum distillation.
\newblock In {\em NeurIPS}, pages 9694--9705, 2021.

\bibitem{li2020hero}
Linjie Li, Yen-Chun Chen, Yu~Cheng, Zhe Gan, Licheng Yu, and Jingjing Liu.
\newblock {HERO}: Hierarchical encoder for video+ language omni-representation
  pre-training.
\newblock {\em EMNLP}, pages 2046--2065, 2020.

\bibitem{lin2014microsoft}
Tsung-Yi Lin, Michael Maire, Serge Belongie, James Hays, Pietro Perona, Deva
  Ramanan, Piotr Doll{\'a}r, and C.~Lawrence Zitnick.
\newblock Microsoft {COCO}: Common objects in context.
\newblock In {\em ECCV}, pages 740--755, 2014.

\bibitem{liu2021hit}
Song Liu, Haoqi Fan, Shengsheng Qian, Yiru Chen, Wenkui Ding, and Zhongyuan
  Wang.
\newblock {HiT}: Hierarchical transformer with momentum contrast for video-text
  retrieval.
\newblock In {\em ICCV}, pages 11915--11925, 2021.

\bibitem{liu2019use}
Yang Liu, Samuel Albanie, Arsha Nagrani, and Andrew Zisserman.
\newblock Use what you have: Video retrieval using representations from
  collaborative experts.
\newblock In {\em BMVC}, page 279, 2019.

\bibitem{LoshchilovH19}
Ilya Loshchilov and Frank Hutter.
\newblock Decoupled weight decay regularization.
\newblock In {\em ICLR}, 2019.

\bibitem{lu2022cots}
Haoyu Lu, Nanyi Fei, Yuqi Huo, Yizhao Gao, Zhiwu Lu, and Ji-Rong Wen.
\newblock {COTS}: Collaborative two-stream vision-language pre-training model
  for cross-modal retrieval.
\newblock In {\em CVPR}, pages 15692--15701, 2022.

\bibitem{luo2020univilm}
Huaishao Luo, Lei Ji, Botian Shi, Haoyang Huang, Nan Duan, Tianrui Li, Xilin
  Chen, and Ming Zhou.
\newblock {UniVL}: A unified video and language pre-training model for
  multimodal understanding and generation.
\newblock {\em arXiv preprint arXiv:2002.06353}, 2020.

\bibitem{luo2021clip4clip}
Huaishao Luo, Lei Ji, Ming Zhong, Yang Chen, Wen Lei, Nan Duan, and Tianrui Li.
\newblock {CLIP4Clip}: An empirical study of clip for end to end video clip
  retrieval.
\newblock {\em arXiv preprint arXiv:2104.08860}, 2021.

\bibitem{miech2020end}
Antoine Miech, Jean-Baptiste Alayrac, Lucas Smaira, Ivan Laptev, Josef Sivic,
  and Andrew Zisserman.
\newblock End-to-end learning of visual representations from uncurated
  instructional videos.
\newblock In {\em CVPR}, pages 9879--9889, 2020.

\bibitem{miech2019howto100m}
Antoine Miech, Dimitri Zhukov, Jean-Baptiste Alayrac, Makarand Tapaswi, Ivan
  Laptev, and Josef Sivic.
\newblock {HowTo100M}: Learning a text-video embedding by watching hundred
  million narrated video clips.
\newblock In {\em ICCV}, pages 2630--2640, 2019.

\bibitem{mithun2018learning}
Niluthpol~Chowdhury Mithun, Juncheng Li, Florian Metze, and Amit~K
  Roy-Chowdhury.
\newblock Learning joint embedding with multimodal cues for cross-modal
  video-text retrieval.
\newblock In {\em ICMR}, pages 19--27, 2018.

\bibitem{oord2018representation}
Aaron van~den Oord, Yazhe Li, and Oriol Vinyals.
\newblock Representation learning with contrastive predictive coding.
\newblock {\em arXiv preprint arXiv:1807.03748}, 2018.

\bibitem{ordonez2011im2text}
Vicente Ordonez, Girish Kulkarni, and Tamara Berg.
\newblock {Im2Text}: Describing images using 1 million captioned photographs.
\newblock In {\em NeurIPS}, pages 1143--1151, 2011.

\bibitem{patrick2020support}
Mandela Patrick, Po{-}Yao Huang, Yuki~Markus Asano, Florian Metze, Alexander~G.
  Hauptmann, Jo{\~{a}}o~F. Henriques, and Andrea Vedaldi.
\newblock Support-set bottlenecks for video-text representation learning.
\newblock In {\em ICLR}, 2021.

\bibitem{radford2021learning}
Alec Radford, Jong~Wook Kim, Chris Hallacy, Aditya Ramesh, Gabriel Goh,
  Sandhini Agarwal, Girish Sastry, Amanda Askell, Pamela Mishkin, Jack Clark,
  Gretchen Krueger, and Ilya Sutskever.
\newblock Learning transferable visual models from natural language
  supervision.
\newblock In {\em ICML}, pages 8748--8763, 2021.

\bibitem{sharma2018conceptual}
Piyush Sharma, Nan Ding, Sebastian Goodman, and Radu Soricut.
\newblock Conceptual captions: A cleaned, hypernymed, image alt-text dataset
  for automatic image captioning.
\newblock In {\em ACL}, pages 2556--2565, 2018.

\bibitem{TanB19}
Hao Tan and Mohit Bansal.
\newblock {LXMERT:} learning cross-modality encoder representations from
  transformers.
\newblock In Kentaro Inui, Jing Jiang, Vincent Ng, and Xiaojun Wan, editors,
  {\em EMNLP-IJCNLP}, pages 5099--5110, 2019.

\bibitem{vaswani2017transformer}
Ashish Vaswani, Noam Shazeer, Niki Parmar, Jakob Uszkoreit, Llion Jones,
  Aidan~N. Gomez, Lukasz Kaiser, and Illia Polosukhin.
\newblock Attention is all you need.
\newblock In {\em NeurIPS}, pages 5998--6008, 2017.

\bibitem{venugopalan2014translating}
Subhashini Venugopalan, Huijuan Xu, Jeff Donahue, Marcus Rohrbach, Raymond~J.
  Mooney, and Kate Saenko.
\newblock Translating videos to natural language using deep recurrent neural
  networks.
\newblock In {\em NAACL-HLT}, pages 1494--1504, 2015.

\bibitem{wang2021t2vlad}
Xiaohan Wang, Linchao Zhu, and Yi~Yang.
\newblock {T}2{VLAD}: global-local sequence alignment for text-video retrieval.
\newblock In {\em CVPR}, pages 5079--5088, 2021.

\bibitem{wang2021vatex}
Xin Wang, Jiawei Wu, Junkun Chen, Lei Li, Yuan{-}Fang Wang, and William~Yang
  Wang.
\newblock {VaTeX}: {A} large-scale, high-quality multilingual dataset for
  video-and-language research.
\newblock In {\em ICCV}, pages 4580--4590, 2019.

\bibitem{wu2018unsupervised}
Zhirong Wu, Yuanjun Xiong, Stella~X Yu, and Dahua Lin.
\newblock Unsupervised feature learning via non-parametric instance
  discrimination.
\newblock In {\em CVPR}, pages 3733--3742, 2018.

\bibitem{xie2018rethinking}
Saining Xie, Chen Sun, Jonathan Huang, Zhuowen Tu, and Kevin Murphy.
\newblock Rethinking spatiotemporal feature learning: Speed-accuracy trade-offs
  in video classification.
\newblock In {\em ECCV}, pages 318--335, 2018.

\bibitem{XuZX0Z0Z17}
Dejing Xu, Zhou Zhao, Jun Xiao, Fei Wu, Hanwang Zhang, Xiangnan He, and Yueting
  Zhuang.
\newblock Video question answering via gradually refined attention over
  appearance and motion.
\newblock In {\em ACMMM}, pages 1645--1653, 2017.

\bibitem{xu2016msr}
Jun Xu, Tao Mei, Ting Yao, and Yong Rui.
\newblock {MSR-VTT}: A large video description dataset for bridging video and
  language.
\newblock In {\em CVPR}, pages 5288--5296, 2016.

\bibitem{YangMSLS21}
Antoine Yang, Antoine Miech, Josef Sivic, Ivan Laptev, and Cordelia Schmid.
\newblock Just ask: Learning to answer questions from millions of narrated
  videos.
\newblock In {\em ICCV}, pages 1666--1677, 2021.

\bibitem{yang2021taco}
Jianwei Yang, Yonatan Bisk, and Jianfeng Gao.
\newblock {TACo}: Token-aware cascade contrastive learning for video-text
  alignment.
\newblock In {\em ICCV}, pages 11562--11572, 2021.

\bibitem{zhang2018cross}
Bowen Zhang, Hexiang Hu, and Fei Sha.
\newblock Cross-modal and hierarchical modeling of video and text.
\newblock In {\em ECCV}, pages 385--401, 2018.

\end{thebibliography}
}

\appendix

\section{Appendix}

\vspace{-0.2cm}
\subsection{Limitations and Potential Negative Societal Impacts}
\vspace{-0.1cm}

\noindent\textbf{Limitations.}~~
The key idea of our LGDN is to propose the SFP mechanism to filter out noisy/redundant frames for fine-grained semantic alignment, along with MVCL for capturing global temporal information. In most downstream tasks, these two modules are complementary to each other. And we also observe that only a few salient frames (e.g., 2 ones) are enough for most downstream tasks, and thus we do not consider aggregating temporal information across salient frames.
However, the SFP mechanism may need to be slightly changed when facing specified scenarios (e.g., long-term complicated videos over 30 minutes that highly rely on temporal information). 
On the one hand, we could adjust the weights between the two modules (MVCL and SFP) according to the situation. On the other hand, we can split the full video into several clips (e.g., 3 minutes per clip), apply our SFP mechanism on each clip, and obtain the salient frames from all clips. In this way, we could consider aggregating temporal information across salient frames.

\noindent\textbf{Potential Negative Societal Impacts.}~~
Video-language learning, especially large-scale video-language modeling, has developed rapidly over the past few years and led to the greatest advance in search engines, video recommendation, and multimedia data management. Despite its effectiveness, existing video-language pre-training models still face possible risks. As these models often rely on a large amount of web data, they may acquire biases or prejudices (especially in search engines and recommendation systems), which must be properly addressed before model deploying.

\begin{figure}[t]
    \centering
    \includegraphics[width=0.85\linewidth]{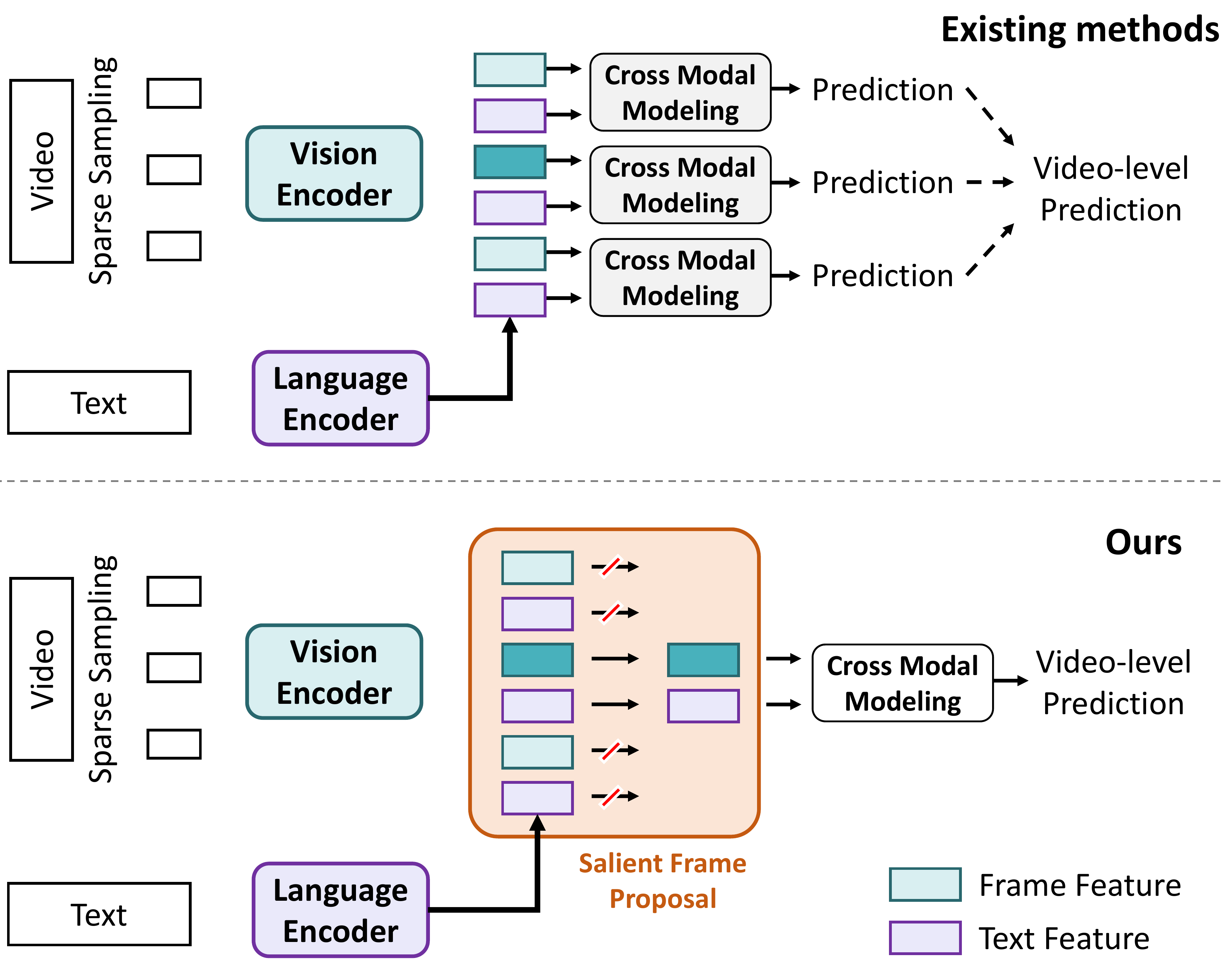}
    \vspace{-0.05in}
    \caption{Comparison between the sparse sampling paradigm (e.g., ClipBERT, top) and our proposed LGDN (bottom). Different from most existing methods that utilize all extracted video features, LGDN dynamically filters out the unmatched or redundant frames under the language supervision. }
    \label{fig:sample}
    \vspace{-0.1in}
\end{figure}

\vspace{-0.1cm}
\subsection{More Implementation Details}
\vspace{-0.1cm}

\noindent\textbf{Two-Stage Sampling Strategy.}~~ 
The sampling strategy of our LGDN has two stages as shown in Figure~\ref{fig:sample}: 
(1) We first adopt sparse sampling to sample 16 frames from each video before feeding them into the LGDN, which is the same as ClipBERT. 
(2) We further utilize salient sampling (SFP) to select a few salient frames (from 16 frames per video) before fusion layers. 

\noindent\textbf{Details of Network Architecture.}~~ 
We adopt ViT-B/16~\cite{alexey2021vit} as our frame encoder and the first 6 layers of BERT-base~\cite{jacob2019bert} as our text encoder.
The dimensions of the output vectors of the frame and text tokens are both $N_{seq}\times768$, where $N_{seq}$ is the sequence length. For each frame/text, the final output vector of the [CLS] token is used as the frame/text embedding. 
We utilize one Transformer layer (with 768 hidden units and 12 heads) as the temporal module to aggregate the frame embeddings to obtain the video embedding. 
We then utilize a single fully-connected layer for each modality to project the frame/video/text embeddings to the joint cross-modal space. The final dimensions of the frame and text embeddings are 256.
We further apply a 6-layer cross-attention Transformer with an additional cross-attention module in each layer as our multi-modal encoder as in LXMERT~\cite{TanB19} and ALBEF~\cite{li2021albef}, where the network parameters are initialized from the last 6 layers of BERT-base. Each layer of the cross-attention Transformer consists of a self-attention sub-layer, a cross-attention sub-layer, and a feed-forward sub-layer with 768 hidden units and 12 heads. 

\noindent\textbf{Downstream Datasets.}~~
(1) \textbf{MSR-VTT}~\cite{xu2016msr} is a popular video-text dataset with three data partitions. The full split is the official partition which uses 6,513/497/2,990 videos for training/validation/testing. The 1k-A split is a widely-used partition with 9,000/1,000 videos for training/testing. Recent works also apply the 7k-1k split, which uses 7,000/1,000 videos for training/testing. All three data partitions are considered in our experiments. 
(2) \textbf{MSVD}~\cite{xu2016msr} contains 80K descriptions for 1,970 videos from YouTube. Following Frozen in Time~\cite{bain2021frozen}, we employ the standard split with 1,200 videos for training and 670 videos for testing.
(3) \textbf{DiDeMo}~\cite{anne2017localizing} consists of 10K videos and 40K sentences. Each sentence includes the temporal localization information. Following Frozen in Time~\cite{bain2021frozen}, we conduct the paragraph-to-video retrieval task, where all descriptions in the same video are concatenated into a single description. 
(4) \textbf{VATEX}~\cite{wang2021vatex} is composed of 34,911 videos. We use 25,991/1,500/1,500 videos for training/validation/testing, following the split in Support Set~\cite{patrick2020support}.
(5) \textbf{MSRVTT-QA}~\cite{XuZX0Z0Z17} is a widely-used video-question answering dataset. We employ the standard split as in ClipBERT~\cite{lei2021less}.

\noindent\textbf{Resources Used.}~~
It takes around 3 / 9 days to pre-train LGDN (5.2M / 15.2M) with 16 Tesla V100 GPUs. For each downstream task, it takes about 5-15 hours with 8 Tesla V100 GPUs. 

\vspace{-0.1cm}
\subsection{More Experimental Results}
\vspace{-0.1cm}

\begin{figure}[ht!]
    \centering
    \includegraphics[width=0.99\textwidth]{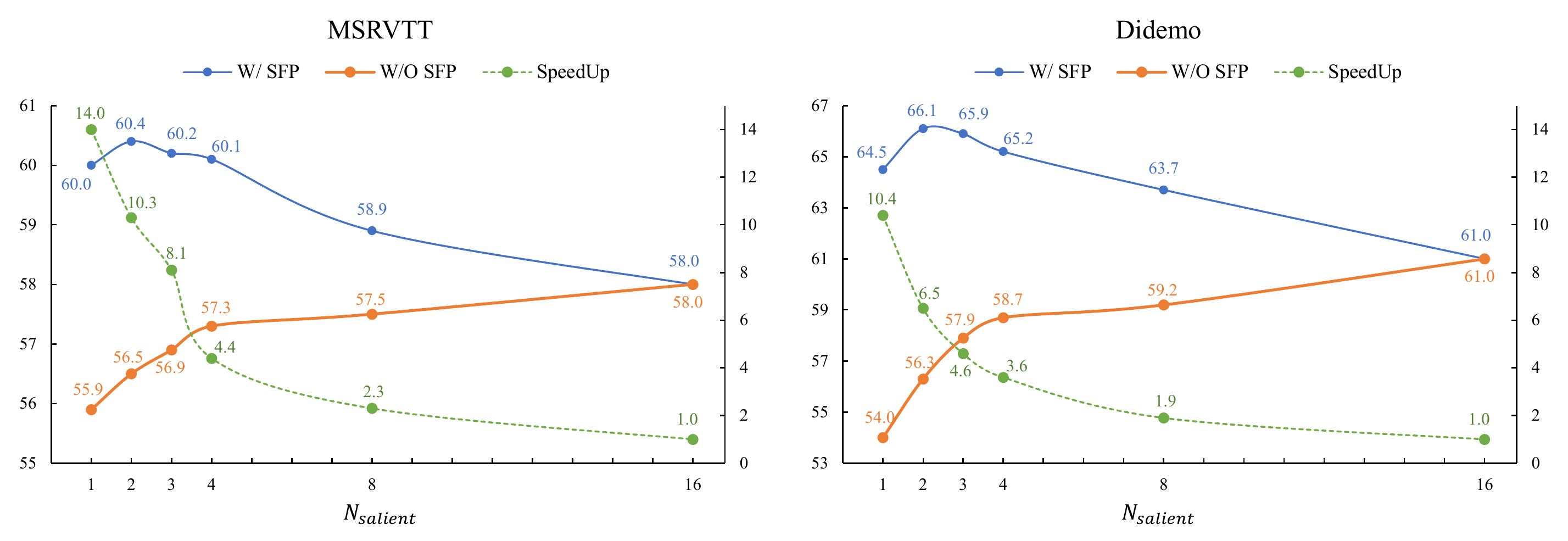}
    \vspace{-0.1in}
    \caption{
    Effect of value change of $N_{salient}$ on the performance of our LGDN for text-to-video retrieval on the MSR-VTT / Didemo test set. w/ SFP: frames are filtered by our SFP mechanism. w/o SFP: frames are randomly selected. SpeedUp: the speedup over $N_{salient}=16$. 
    }
    \label{fig:abl_msrvtt_appendix}
\end{figure}

\noindent\textbf{Effect of SFP mechanism.}~~
To further demonstrating the effectiveness of SFP mechanism, 
we conduct experiments considering model variants that select $N_{salient}$ salient frames from $N = 16$ frames by our SFP or just randomly select $N_{salient}$ frames (denoted as w/o SFP) for token-level alignment in Figure~\ref{fig:abl_msrvtt_appendix}. 
We sample $N = 16$ frames from each video and evaluate different model variants that use $N_{salient} \in \{1, 2, 3, 4, 8, 16\}$ frames for token-level alignment. Note that when $N_{salient} = 16$, sampling by our SFP degrades to w/o SFP. As expected, when randomly selecting frames like most existing methods, adding more frames does bring better results. However, utilizing only $N_{salient} = \{2, 3, 4\}$ salient frames filtered by our SFP significantly outperforms random sampling (i.e., w/o SFP), and even outperforms utilizing all 16 extracted frames. This suggests that our SFP mechanism not only selects correct salient frames but also alleviates the noise problem. 

\begin{figure}[ht!]
    \centering
    \includegraphics[width=0.99\textwidth]{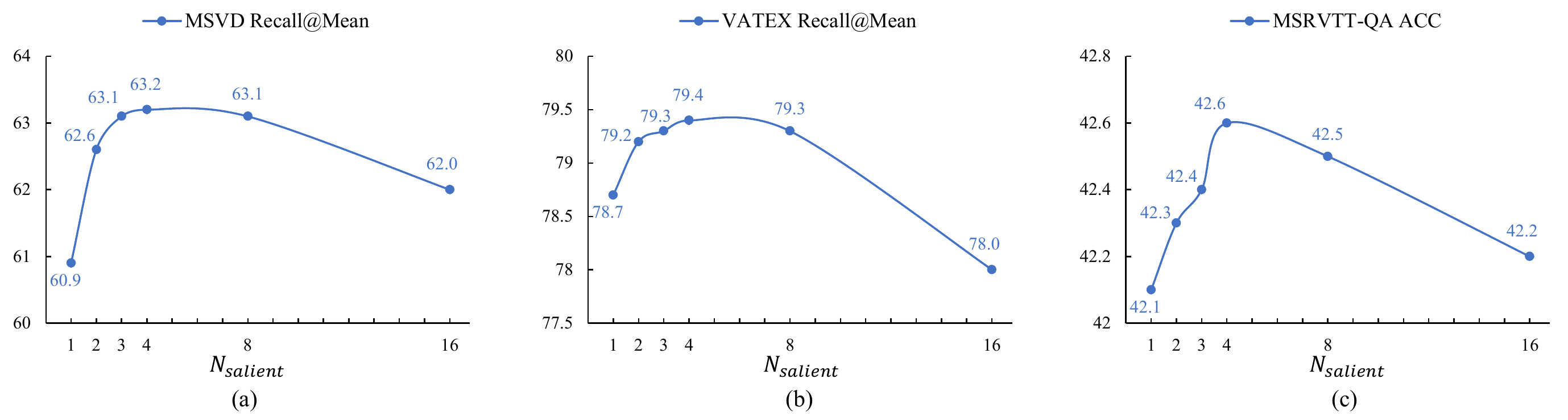}
    \vspace{-0.1in}
    \caption{
    (a) Effect of value change of $N_{salient}$ on the performance of our LGDN for text-to-video retrieval on the MSVD test set. (b) Effect of value change of $N_{salient}$ on the performance of our LGDN for text-to-video retrieval on the VATEX test set. (c) Effect of value change of $N_{salient}$ on the performance of our LGDN for video question answering on the MSRVTT-QA val set.
    }
    \label{fig:abl2_frame_appendix}
    \vspace{0.05in}
\end{figure}

We also present the experiment results on other public datasets in Figure~\ref{fig:abl2_frame_appendix}. Though the best parameters $N_{salient}$ are not quite the same on different datasets ($N_{salient} = 2$ for MSRVTT, Dedimo; $N_{salient} = 4$ for MSVD, VATEX, MSRVTT-QA), it can be observed that SFP mechanism  significantly improves the baseline. Meanwhile, the performance changes among the three parameters ($N_{salient} \in \{2, 3, 4\}$) are very marginal, which further verifies the robustness and effectiveness of our LGDN. 

\begin{table}[t!]
    \centering
    \caption{Effect of value change of memory bank size on the performance of our LGDN model. Video-text retrieval results are reported on the MSR-VTT 1k-A test set. }
    \vspace{-0.05in}
    \scalebox{0.93}{
    \tabcolsep16pt
    \begin{tabular}{cccccccc}
    \toprule
    \multirow{2}{*}{Bank Size}  & \multicolumn{3}{c}{Text-to-Video Retrieval} & \multicolumn{3}{c}{Video-to-Text Retrieval}  \\
     & R@1 & R@5  & MdR & R@1 & R@5 & MdR  \\
    \midrule
     1,200 & 36.0 & 65.2 & 3.0 & 36.2 & 64.8 & 3.0 \\
    2,400 &  36.9 & 64.6 & 3.0 & 37.0 & 64.8 & \bf2.0 \\
    4,800 &  37.7  & \bf65.7 & 3.0 & 37.8 & 65.0 & \bf2.0 \\
    9,600 & \bf38.9 & \bf65.7 & \bf2.0 & \bf37.9 & \bf65.4 & \bf2.0\\
    19,200 &  38.3 & 64.9 & 3.0 & 37.5 & 64.3 & 3.0 \\
    \bottomrule
    \end{tabular}}
    \label{tab:queue_size}
    \vspace{-0.0in}
\end{table}

\noindent\textbf{Effect of Value Change of Memory Bank Size.}~~
In Table~\ref{tab:queue_size}, we show the influence of different values of the memory bank size $N_m$ on the performance of our LGDN. With the increase of memory bank size $N_m$, the performance of our LGDN begins with an increase, indicating that the introduction of large-scale negatives for contrastive learning indeed brings performance improvements. However, when $N_m$ becomes too large ($>9,600$), the performance drops slightly. One possible reason is that too large memory bank size may introduce more hard negative samples, which makes it harder to learn a good vision-language representation. Our LGDN thus performs the best at $N_m=9,600$. 

\begin{table}[t!]
\centering
    \caption{The speed and memory cost of different sampling strategies on the Deidemo test set.}
    \label{tab:memory}
    \vspace{-0.08in}
    \scalebox{0.95}{
        \tabcolsep20pt
        \begin{tabular}{@{}l|ccc@{}}
        \toprule
        Sampling Strategy &  Speedup    & Memory Cost & R@SUM \\
        \midrule
        Sparse sampling    & 1.0x        & 1.0x  & 183.0 \\
        \begin{tabular}[c]{@{}c@{}}Salient sampling ($N_{salient}=1$)\end{tabular} & 10.4x       & 0.60x & 193.5 \\
        \begin{tabular}[c]{@{}c@{}}Salient sampling ($N_{salient}=2$)\end{tabular} & 6.5x        & 0.62x & 198.3 \\
        \begin{tabular}[c]{@{}c@{}}Salient sampling ($N_{salient}=4$)\end{tabular} & 3.6x        & 0.68x & 195.6 \\
        \bottomrule
    \end{tabular}}
\end{table}

\begin{figure}[t!]
    \centering
    \includegraphics[width=0.98\textwidth]{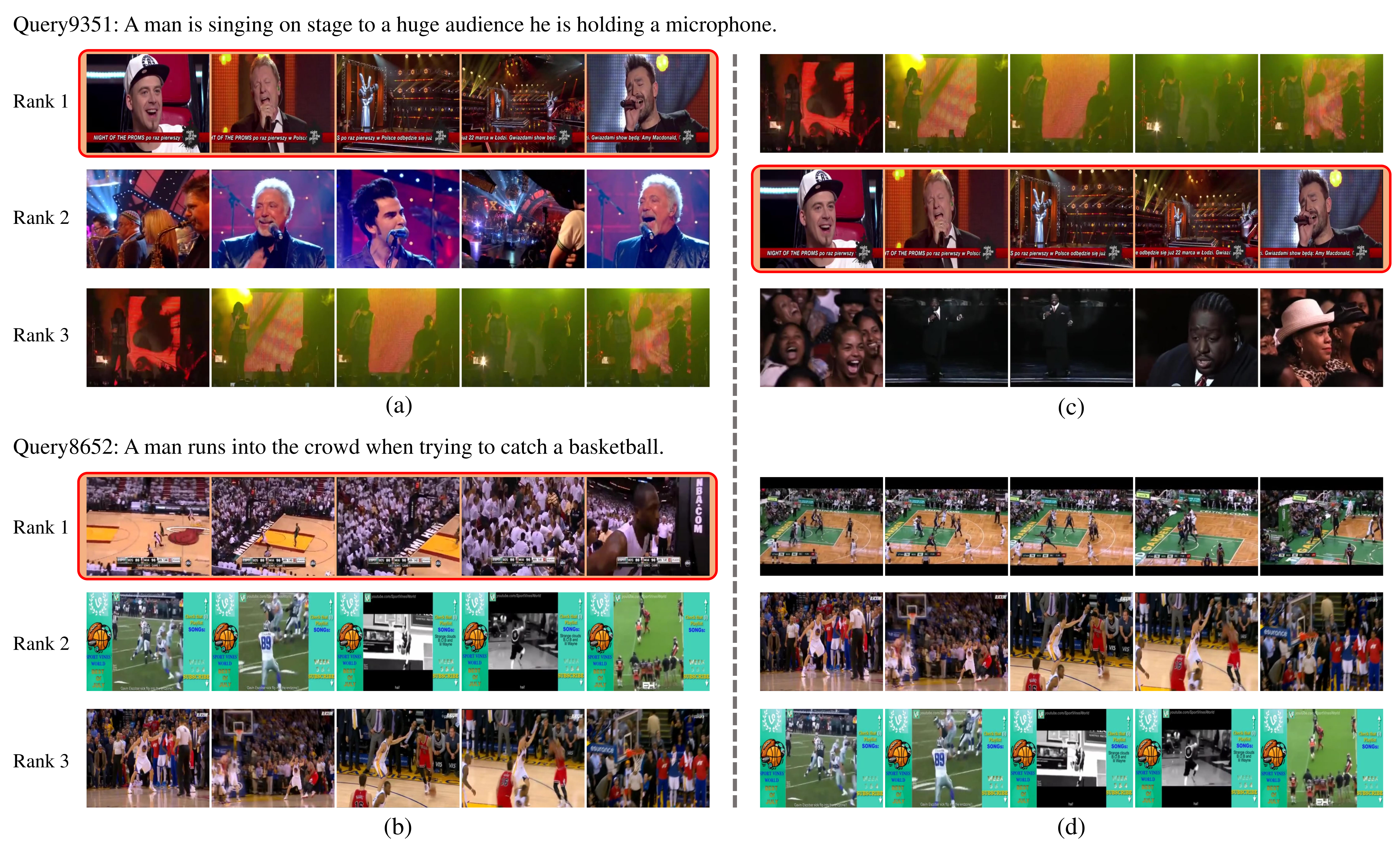}
    \vspace{-0.1in}
    \caption{
    Top-3 text-to-video retrieval results on the MSR-VTT 1k-A test set (correct result is highlighted in red). The left are videos ranked by our LGDN, while the right are results from our model without using the SFP mechanism.
    }
    \label{fig:visualize_appendix2}
    \vspace{-0.1in}
\end{figure}

\noindent\textbf{Speed and Memory Cost.}~~
We present the speed and memory cost on the Didemo test set in Table~\ref{tab:memory}. For fair comparison, all experiments are conducted on 8 Tesla-V100 GPUs with mini-batch size 24. It can be seen that our salient sampling strategy is obviously faster and costs less memory, as compared with sparse sampling (utilizing all 16 frames for feature extraction and multi-modal fusion).

\begin{figure*}[t]
    \centering
    \includegraphics[width=0.96\textwidth]{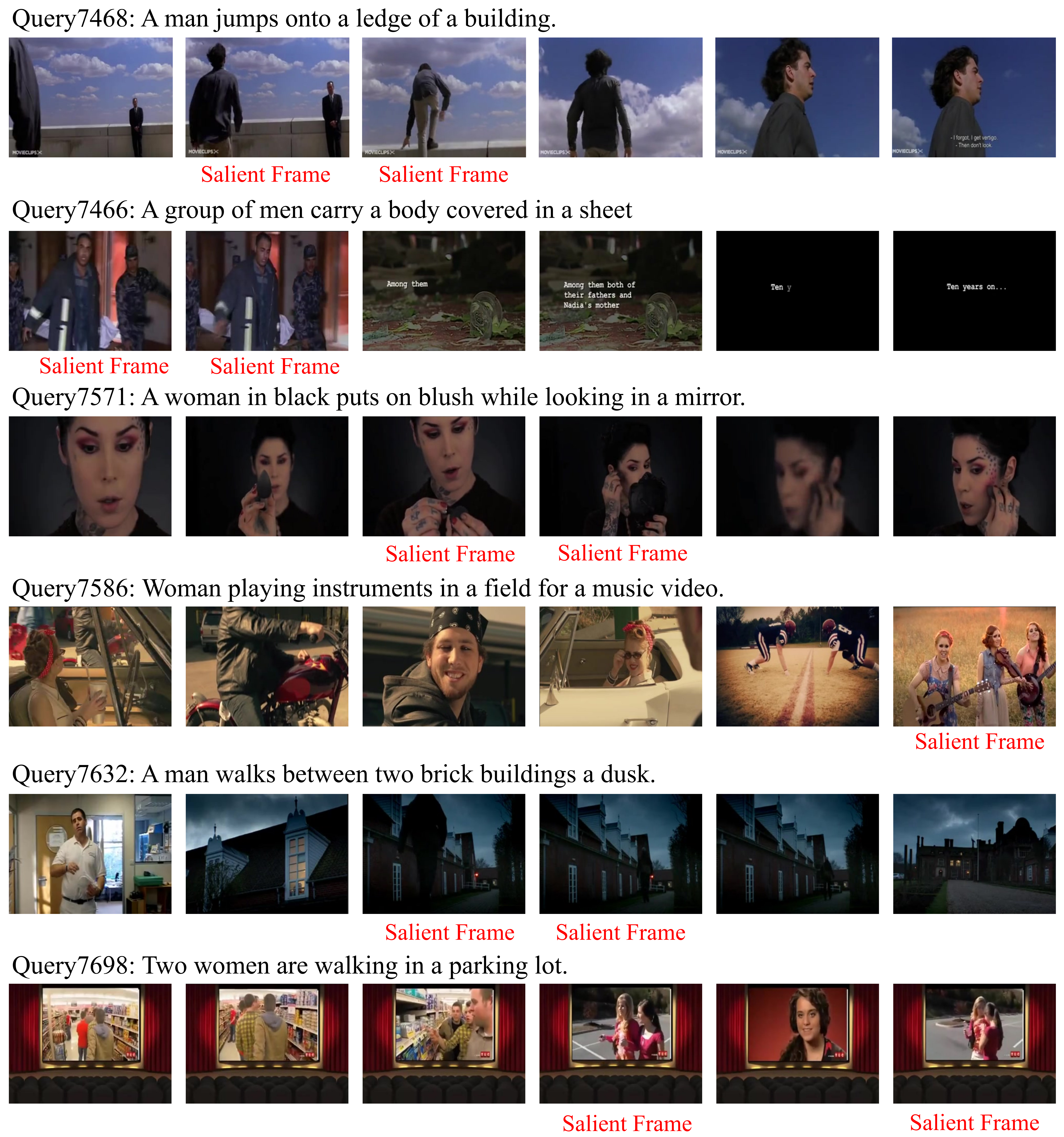}
    \vspace{-0.05in}
    \caption{
    Visualization of salient frames obtained by our LGDN on the MSR-VTT test set. We uniformly sample 6 frames from each video, among which the red ones denote the salient frames selected by our SFP mechanism. We find that our SFP mechanism indeed filters out the unmatched/redundant frames under the language supervision.
    }
    \label{fig:visualize_appendix1}
    \vspace{-0.1in}
\end{figure*}

\vspace{-0.1cm}
\subsection{Visualization Results}
\vspace{-0.1cm}

We provide visualization examples in Figure~\ref{fig:visualize_appendix2}. Figures~\ref{fig:visualize_appendix2}(a)-(b) are the retrieved results by our LGDN and Figures~\ref{fig:visualize_appendix2}(c)-(d) are the retrieved results  by our model without using the SFP mechanism. We can see from Figures~\ref{fig:visualize_appendix2}(a)-(b) that although the target videos have noisy frames (e.g., in the first frame of rank 1 video of Figure~\ref{fig:visualize_appendix2}(a), the man is laughing; the last frame of rank 1 video of Figure~\ref{fig:visualize_appendix2}(b) is the close up of the player who run into the crowd), LGDN precisely retrieves the ground-truth. In Figure~\ref{fig:visualize_appendix2}(c), LGDN without using SFP also retrieves the corresponding video where the people are singing to the audience, however, in rank 1 video, there are a woman and a man both singing, which is incorrect to the query ``A man''. In Figure~\ref{fig:visualize_appendix2}(d), LGDN without using SFP only retrieves video corresponding to basketball and crowd. These examples indicate that the noisy information misleads cross-modal modeling and our LGDN with SFP can help to alleviate it. 

\begin{figure*}[t]
    \centering
    \includegraphics[width=0.96\textwidth]{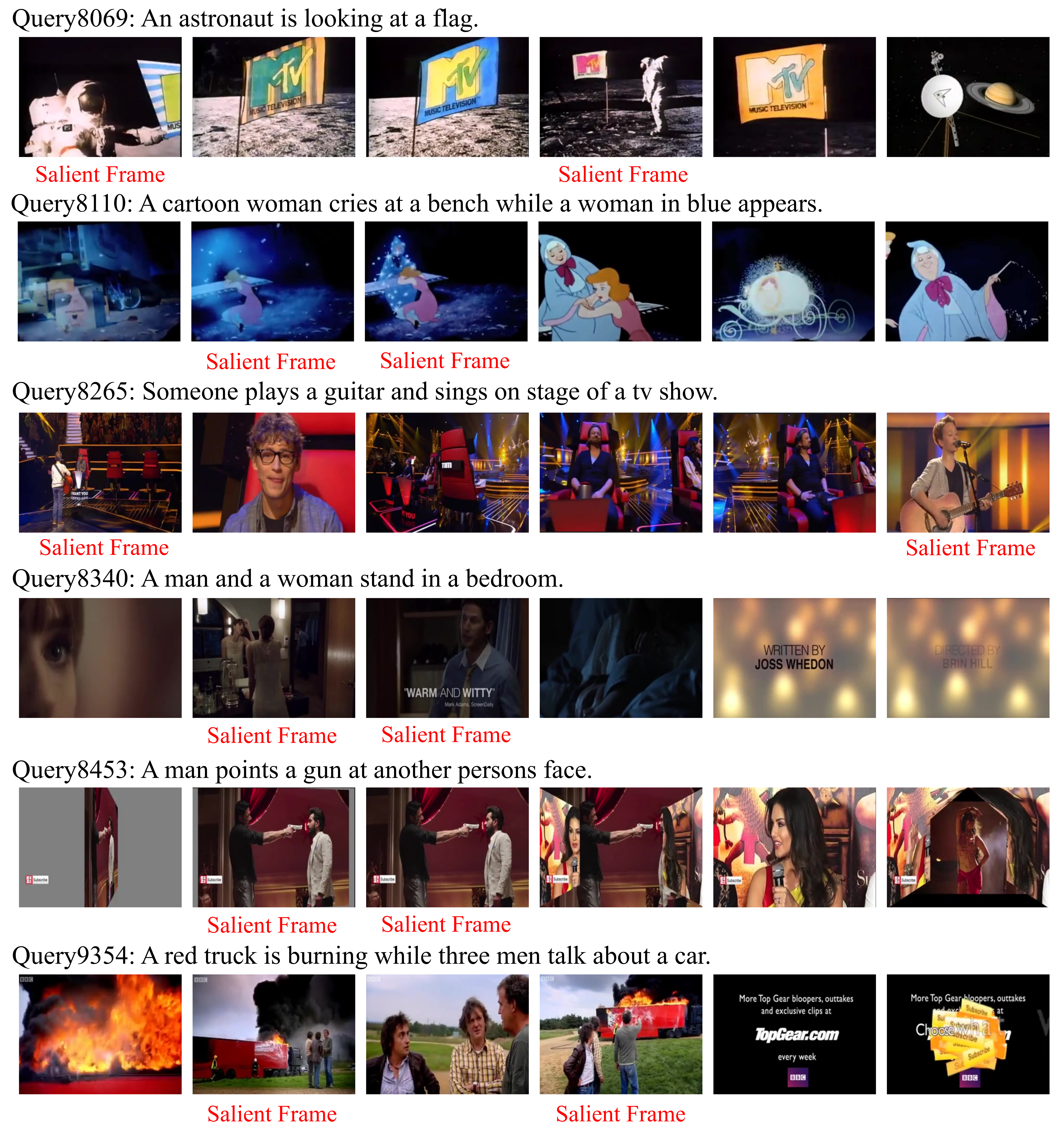}
    \vspace{-0.05in}
    \caption{
    More visualization of salient frames obtained by our LGDN on the MSR-VTT test set.
    }
    \label{fig:visualize_appendix3}
    \vspace{-0.1in}
\end{figure*}

Further, we provide more visualization results obtained by our LGDN in Figures~\ref{fig:visualize_appendix1}-\ref{fig:visualize_appendix3}. We uniformly sample 6 frames from each video, among which the red ones denote salient frames selected by the SFP mechanism. It can be clearly observed that: (1) Although the holistic video is semantically related to the paired text, there still exist noisy frames (e.g., the transition in Frame 3-6 of Query7466) and unrelated frames (e.g., Frame 4-6 of Query7468, Frame 1-5 of Query7586, Frame 2-3 of Query8069, and Frame 2-5 of Query8265). (2) The salient frames obtained from the SFP mechanism correctly represent the semantic information of the video given the paired text, which indeed helps our LGDN to precisely filter out noisy information for better video-language modeling.

\end{document}